\documentclass{article}

\usepackage[utf8]{inputenc}
\usepackage[T1]{fontenc}
\usepackage[letterpaper, margin=1in]{geometry} 

\usepackage{natbib} 
\usepackage{stfloats}
\usepackage{mathrsfs}
\usepackage{amsmath,amssymb}
\usepackage{graphicx,subfigure}
\usepackage{algorithm,algorithmic} 
\usepackage{longtable,booktabs}
\usepackage{enumitem}
\usepackage{placeins}
\usepackage{float}
\usepackage{url}
\usepackage{bm} 

\usepackage[printonlyused]{acronym}
\acrodef{PDF}{probability density function}
\acrodef{PSO}{Particle Swarm Optimization}
\acrodef{CPS}{Constriction-based Particle Swarm Optimizer}
\acrodef{IPS}{Inertia-weighted Particle Swarm Optimizer}
\acrodef{ETO}{Exponential Trigonometric Optimization}
\acrodef{AOA}{Arithmetic Optimization Algorithm}
\acrodef{GJO}{Golden Jackal Optimization}
\acrodef{GWO}{Grey Wolf Optimizer}
\acrodef{HGS}{Hunger Games Search}
\acrodef{HHO}{Harris Hawks Optimization}
\acrodef{SCA}{Sine Cosine Algorithm}
\acrodef{SCH}{Sinh CosH optimizer}

\usepackage[all]{nowidow}
\def\newblock{\hskip .11em plus .33em minus .07em} 

\usepackage{amsthm}
\theoremstyle{plain} 

\theoremstyle{definition} 

\theoremstyle{remark} 



\begin{document}%
	
	\title{\textbf{On the Structural and Statistical Flaws of the Exponential-Trigonometric Optimizer}}
	
	\author{Ngaiming Kwok \\ 		Independent Research	}
	\date{}
	
	\maketitle
	
	\begin{abstract}
		The proliferation of metaphor-based metaheuristics has often been accompanied by issues of symbolic inflation, benchmarking opacity, and statistical misuse. This study presents a diagnostic critique of the recently proposed Exponential Trigonometric Optimizer (ETO), exposing fundamental flaws in its algorithmic structure and the statistical reporting of its performance. Through a stripped mathematical reconstruction, we identify inert symbolic constructs, ill-defined recurrence schedules, and ineffective update mechanisms that collectively undermine the algorithm's purported balance and effectiveness. A principled benchmarking comparison against nine established metaheuristics on the CEC 2017 and 2021 suites reveals that ETO's performance claims are inflated. While it demonstrates mid-tier competitiveness, it consistently fails against top-tier algorithms, especially under high-dimensional and shift-rotated landscapes. Our statistical framework, employing rank-based non-parametric tests and effect size diagnostics, quantifies these limitations and highlights ETO's structural fragility and lack of scalability. The paper concludes by advocating for a reformist framework in metaheuristic research, emphasizing symbolic hygiene, operator attribution, and statistical transparency to mitigate misleading narratives and foster a more robust and reproducible optimization literature.
	\end{abstract}
	


	\section{Introduction}\label{sec:introduction}
	
	Metaheuristics have evolved over five decades into a prolific and diverse class of optimization algorithms, with applications spanning engineering, logistics, and artificial intelligence. As reviewed by Mart{\'\i} et al. \cite{marti2025fifty}, the field has witnessed exponential growth in algorithmic proposals, many of which are metaphor-based and inspired by natural or social phenomena. While this expansion reflects creative enthusiasm, it has also led to a proliferation of heuristics with questionable novelty, inconsistent benchmarking, and limited reproducibility.
	
	S{\"o}rensen \cite{sorensen2015metaheuristics} critically exposed the metaphor-centric culture of metaheuristic design, arguing that metaphors often obscure algorithmic mechanisms and hinder scientific scrutiny. This concern was echoed by Aranha et al. \cite{aranha2022metaphor}, who called for principled evaluation, reproducible experimentation, and editorial reform to address the symbolic inflation and benchmarking opacity that plague the literature. Despite these calls, many recent publications continue to over-claim performance without sufficient diagnostic evidence or replicable methodology.
	
	This study responds to that call for action by conducting a diagnostic benchmarking comparison of ten metaheuristic algorithms, with a particular focus on evaluating the performance claims made in the \ac{ETO} paper \cite{luan2024exponential}. Although the source code for \ac{ETO} is publicly available, the original study lacks parameter transparency and statistical rigour. Through controlled experimentation on the CEC 2021 and CEC 2017 benchmark suites, we expose symbolic inflation, rank instability, and convergence stagnation in \ac{ETO}'s reported results.
	
	To ensure principled and reproducible evaluation, we assess all algorithms under standardized experimental conditions and apply rank-based non-parametric statistics. Our framework integrates the Friedman test for global rank variance, the Dunn–Šidák corrected Wilcoxon signed-rank test for pairwise comparisons, and effect size metrics such as Cliff’s $\delta$ and rank correlation $r$ to quantify performance gaps. Quartile stratification and median difference analysis further support symbolic tagging and convergence diagnostics. These procedures address common statistical deficiencies in the literature—such as reliance on mean-based comparisons without dispersion or effect sizes, and the use of uncorrected pairwise tests—that obscure variability and inflate claims of superiority.
	
	Such deficiencies are not merely technical oversights; they enable and amplify misleading narratives. In the case of the original \ac{ETO} paper, the misuse of aggregated comparisons and the absence of diagnostic overlays collectively created the illusion of consistent superiority. Without rank-based variance analysis and convergence diagnostics, structural inconsistencies and rank instability remain hidden, allowing symbolic inflation to go undetected. This study exposes how methodological opacity and statistical misuse can be exploited to fabricate algorithmic dominance, reinforcing the urgent need for principled benchmarking and editorial reform.
	
	The key contributions of this study are as follows:
	\begin{enumerate}[nosep]
		\item A principled benchmarking comparison of ten metaheuristics on CEC 2017 and 2021, conducted under standardized and reproducible conditions.  
		\item Reconstruction of the \ac{ETO} algorithm with stripped notation, exposing symbolic redundancy and operator-level inconsistencies.  
		\item Identification of statistical misuses in metaheuristic publishing, and demonstration of corrective practices using non-parametric tests, effect sizes, and quartile-based diagnostics.  
		\item Introduction of a diagnostic framework that integrates symbolic hygiene, operator attribution, and statistical transparency to guide reproducible evaluation.  
		\item Advocacy for editorial reform through methodological practices that improve benchmarking reliability and mitigate symbolic inflation.  
	\end{enumerate}
	
	The remainder of the paper is structured as follows. Section \ref{sec:recollection} reconstructs the \ac{ETO} algorithm using stripped notation, exposing its symbolic structure and control flow, and identifying flaws in operator design, randomness, and structural logic. Section \ref{sec:comparaive} presents a comparative study across CEC 2017 and 2021 benchmarks. Section \ref{subsec:convergence} analyses convergence behaviours with stratified overlays and inflation diagnostics, while Section \ref{subsec:statistical} reports statistical tests to validate or challenge visual impressions. Section \ref{sec:conclusion} concludes with future directions for principled benchmarking.

	\section{Recollection and Diagnosis}\label{sec:recollection}
	
	We provide a structured recollection of the \ac{ETO} algorithm, followed by a diagnostic analysis of its mathematical formulation. The formulation is presented without metaphor to facilitate interpretability and expose systemic flaws common in metaheuristic literature.
	
	The \ac{ETO} algorithm is inspired by the interplay between exponential decay and trigonometric oscillation, metaphorically likened to dynamic balance in natural systems. This metaphor is used to justify the algorithm’s structure, but it lacks grounding in physical or biological analogues, serving primarily as a narrative device rather than a principled design rationale.
	
	\subsection{Procedures in the \ac{ETO}}
	
	\paragraph{Initialization}
	The optimization process begins by generating an initial population of $N$ agents, each represented by a vector $\mathbf x^0 \in \mathbb{R}^D$. Every agent is initialized uniformly within its prescribed decision interval according to
	\begin{equation}
		\mathbf x^0 = r_0 \cdot (U_b - L_b) + L_b, \quad r_0 \sim \mathscr{U}(0,1)^D,
	\end{equation}
	where $L_b$ and $U_b$ are constant scalar values denoting the lower and upper bounds of the search domain. This ensures uniform and independent sampling across the admissible range of each coordinate. See Eq. (6) in the \ac{ETO} paper.
	
	Following initialization, the objective function is evaluated for all agents, and the current best-so-far position is stored as $\mathbf x_{best}$. The global iteration counter is set to $t = 1$, and any auxiliary coefficients---such as decaying or oscillatory terms---are precomputed to support subsequent stage-dependent dynamics in the algorithm.
	
	\paragraph{Constrained Exploration}
	After the initialization of agents, the iteration proceeds. First, two controls are computed, Eq. (1) and (11) in the \ac{ETO} paper. That is
	\begin{equation}\label{eq:d1_d2}
		d_1 = 0.1 \exp(-0.01t) \cdot \cos(0.5T \cdot (1-t/T )), \qquad d_2 = - d_1.
	\end{equation}
	Then, another control is obtained from, Eq. (18) in the \ac{ETO} paper as,
	\begin{equation}\label{eq:mu_t}
		\mu (t) = 0.01 r_1 \cdot (\sqrt{t/T})^{\tan(d_1/d_2)},\qquad r_1 \sim \mathscr U(0,1).
	\end{equation}
	The control that determines the bounds on the update is given by Eq. (1) and (2) in the \ac{ETO} paper,
	\begin{equation}\label{eq:epsilon}
		\epsilon_i(t) = \lfloor 2-2t \cdot (T - a \cdot \epsilon_i) \rfloor + \epsilon_i, \qquad a=4.6.
	\end{equation}
	where the recursion is started with
	\begin{equation}\label{eq:epsilon_0}
		\epsilon_1 = \lfloor 1+T/b \rfloor, \qquad b=1.55.
	\end{equation}
	
	If the iteration count equals $\epsilon_i(t)$, then the update is constrained between
	\begin{align}\label{eq:UbLb_t}
		U_b^t &= x_{best}^t + r_2 (1 - t/T) \cdot | r_3 x_{best}^t - x^t|,\nonumber \\
		L_b^t &= x_{best}^t - r_2 (1 - t/T) \cdot | r_3 x_{best}^t - x^t|,
	\end{align}
	where $r_2$, $r_3 \sim \mathscr U(0,1)$ are random numbers. See Eq. (3) and (4) in the \ac{ETO} paper.
	
	\paragraph{First Phase of Exploration and Exploitation}
	The transition between the first and second phases is determined by comparing the iteration count to a control parameter computed from Eq. (7) in the \ac{ETO} paper as,
	\begin{equation}
		T_p = \lfloor 1.2 + T/2.25 \rfloor.
	\end{equation}
	If the comparison $t \leq T_p$ is true, then a test on the condition $\mu (t)>1$ is made. If the outcome is true, a control parameter is determined from Eq. (9) in the \ac{ETO} paper,
	\begin{equation}\label{eq:alpha_1}
		\alpha_1 = 3 r_4 (t/T - 0.85) \cdot \exp(d_1/d_2 -1),\qquad r_4 \sim \mathscr U(0,1).
	\end{equation}
	The agent position is updated as
	\begin{equation}\label{eq:update_1}
		x^{t+1} = 
		\begin{cases}
			x_{best}^t + \alpha_1 \cdot |x_{best}^t - x^t|, & r_5 \leq 0.5\\
			x_{best}^t - \alpha_1 \cdot |x_{best}^t - x^t|, & otherwise,
		\end{cases}
	\end{equation}
	where $r_5 \sim \mathscr U(0,1)$ is a random number. See Eq. (8) in the \ac{ETO} paper.
	
	Otherwise, when $\mu (t)>1$ is false, then compute the control parameter, see Eq. (15) in the \ac{ETO} paper,
	\begin{equation}\label{eq:alpha_2}
		\alpha_2 = 3 r_6 (t/T - 0.85) \cdot \exp(|d_1/d_2| - 1.3),\qquad r_6 \sim \mathscr U(0,1).
	\end{equation}
	The agent update is given by Eq. (14) in the \ac{ETO} paper as,
	\begin{equation}\label{eq:update_2}
		x^{t+1} = 
		\begin{cases}
			x_{best}^t + r_7 \alpha_2 \cdot | r_8 x_{best}^t - x^t|, & r_9 < 0.5\\
			x_{best}^t - r_7 \alpha_2 \cdot | r_8 x_{best}^t - x^t|, & otherwise,
		\end{cases}
	\end{equation}
	where $r_7$, $r_8$, $r_9 \sim \mathscr U(0,1)$ are random numbers.
	
	\paragraph{Second Phase of Exploration and Exploitation}
	The second phase is activated when $t > T_p$. Furthermore, when the condition $\mu(t) > 1$ is satisfied, then a further control parameter is computed from Eq. (13) of the \ac{ETO} paper as,
	\begin{equation}\label{eq:alpha_3}
		\alpha_3 = r_{10} \cdot \exp(\tanh(1.5 (-t/T - 0.75) - r_{11})),
	\end{equation}
	where $r_{10}$, $r_{11} \sim \mathscr U(0,1)$. The agent is updated by
	\begin{equation}\label{eq:update_3}
		x^{t+1} = 
		\begin{cases}
			x^t + 3 | r_{12} \alpha_3 x_{best}^t - x^t|, & r_{13} \leq 0.5\\
			x^t - 3 | r_{12} \alpha_3 x_{best}^t - x^t|, & otherwise,
		\end{cases}
	\end{equation}
	where $r_{12}$, $r_{13} \sim \mathscr U(0,1)$. See Eq. (12) in the \ac{ETO} paper.
	
	If $\mu(t) > 1$ returns false, then another control parameter is computed from Eq. (17) in the \ac{ETO} paper as,
	\begin{equation}\label{eq:gamma}
		\gamma = \exp (\tan (d_1/d_2)).
	\end{equation}
	The agent is updated from Eq. (16) in the \ac{ETO} paper,
	\begin{equation}\label{eq:update_4}
		x^{t+1} = x^t + \gamma | r_{14} \alpha_3 x_{best}^t - x^t|,
	\end{equation}
	and $r_{14} \sim \mathscr U(0,1)$ is a random number. After this final procedure the iteration count advances and the procedures repeat until the maximum iteration is reached.
	
	\subsection{Redundant or Inert Symbolic Constructs}
	The ETO algorithm's formulation contains numerous symbolic components that are mathematically inert, functionally redundant, or structurally misaligned with their intended purpose. These provable flaws in the core equations fundamentally undermine the algorithm's purported balance and effectiveness.
	
	\paragraph{Ill-Defined Recurrence Scheduling}
	The update trigger schedule is defined recursively via Eq. \eqref{eq:epsilon} and Eq. \eqref{eq:epsilon_0}. However, this recursive structure is ill-conceived, as the trigger step $\epsilon_i(t)$ increases with the agent index $i$ in a way that rapidly exceeds the maximum number of iterations $T$. To illustrate, let $T = 500$ and adopt $b = 1.55$ as specified in the paper. Then the initial schedule becomes
	\begin{equation}
		\epsilon_i = \lfloor 1 + {500}/{1.55} \rfloor = 323.
	\end{equation}
	For $i = 2$ and $t = 1$, the recursive update yields
	\begin{equation}
		\epsilon_2(1) = \left\lfloor 2 - 2 \cdot (500 - 4.6 \cdot 323) \right\rfloor + 323 = 2296.
	\end{equation}
	This value far exceeds the total iteration budget ($T = 500$), and thus the conditional trigger $t > \epsilon_i(t)$ is never satisfied for any $t \in [1, T]$. As a result, the spatial constraint adaptation defined by Eq. \eqref{eq:UbLb_t} is never invoked, rendering the intended dynamic contraction of the search space entirely inert.
	
	\paragraph{Ineffective Bound Contraction}
	The dynamic update of upper and lower bounds via Eq. \eqref{eq:UbLb_t}, is never actually enforced in any of the subsequent position update rules in Eqs. \eqref{eq:update_1}, \eqref{eq:update_2}, \eqref{eq:update_3} and \eqref{eq:update_4}. No projection, rejection, or sampling constraint ensures that $\mathbf x^t$ remains within $[U_b(t), L_b(t)]$. As such, these expressions are computed but algorithmically ignored. A severe design flaw is thus apparent.
	
	\paragraph{Redundant Oscillatory Coefficients}
	The paired expressions for $d_1$ and $d_2$ are defined in Eq. \eqref{eq:d1_d2}, and are equivalent up to a polarity inversion. Specifically, $d_2 = -d_1$, which implies a fixed ratio $d_1/d_2 = -1$ for all iterations. Consequently, any expressions that depend on this ratio---such as $\tan(d_1/d_2)$ in Eqs. \eqref{eq:mu_t} and \eqref{eq:gamma}, or $\exp(d_1/d_2)$ in Eq. \eqref{eq:alpha_1}---evaluate to constant values: $\tan(-1) = -1.5574$ and $\exp(-1) = 0.3679$. As such, the computation of $d_1$ and $d_2$ introduces symbolic and computation redundancies.
	
	\paragraph{Static Transition Coefficient}
	Equation \eqref{eq:gamma} defines a coefficient intended to modulate step size in the second exploitation phase. Given that $\left|d_1/d_2\right| = 1$, the expression simplifies to $\gamma = \exp(\tan(1)) = 4.7465$, a constant. Furthermore, the claim of ``a delicate combination of exponential and trigonometric functions, creating a unique and effective mechanism.'' cannot be justified with the use of this constant. Most importantly, the derivation of an optimum value for for this constant is not given anywhere in the \ac{ETO} paper.
	
\begin{figure}[h!]
		\centering
		\includegraphics[width=0.35\linewidth]{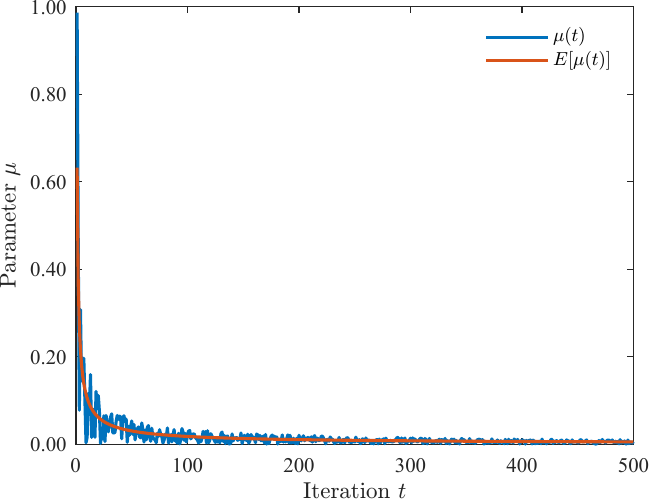}
		\caption{Evolution of control parameter $\mu(t)$ over iterations.}
		\label{fig:Sim_mu_t}
	\end{figure}

	\paragraph{Switching of Agent Updates}
	The update rules for agents are governed by several conditional cases that hinge on the parameter $\mu(t)$, defined in Eq. \eqref{eq:mu_t}. A plot of $\mu(t)$ vs iteration count is shown in Fig. \ref{fig:Sim_mu_t}.
	
	This parameter is compared against a threshold fixed at one. However, since $d_1/d_2 = -1$, we have $\tan(-1) \approx -1.5574$. Taking $T = 500$ and evaluating at $t = 1$, the expression becomes
	\begin{equation}
		\mu(1) = 0.01  r_1 \cdot \left( \sqrt{{1}/{500}} \right)^{-1.5574} \approx 0.01 r_1 \cdot 126.39 = 1.2639  r_1.
	\end{equation}
	The maximum value of $\mu(1)$ (i.e., 1.2639) is only realized when $r_1 = 1$. Since $r_1$ is drawn from a uniform distribution $\mathcal{U}(0,1)$ with expected value $0.5$, the expected value of $\mu(1)$ is approximately $0.6320$. Therefore, the event $\mu(t) > 1$ occurs only with probability
	\begin{equation}
		{P}[\mu(1) > 1] = (1.2639 - 1)/{1.2639} = 0.2088.
	\end{equation}
	At the other extreme, when $t = T=500$, the parameter reduces to $\mu(T) = 0.01 \cdot r_1$. The probability of triggering the exploration condition $\mu(t) > 1$ therefore becomes
	\begin{equation}
		{P}[\mu(500) > 1] = \mathbb{P}[0.01 \cdot r_1 > 1] = {P}[r_1 > 100] = 0,
	\end{equation}
	which is impossible under the uniform distribution $r_1 \sim \mathcal{U}(0,1)$. In other words, the algorithm is deterministically confined to exploitation mode at late iterations. This contradicts the intent of gradual convergence and raises concerns about the validity of the scheduling logic.
	
	Moreover, the \ac{ETO} paper provides no justification for the selection of the threshold value 1, nor does it characterize the expected switching frequency over time. The lack of theoretical support for this stochastic mechanism undermines the rationale for mode adaptation and reflects a broader pattern of symbolic design without analytical grounding.
	
\begin{figure}[h!]
		\centering
		\subfigure[]{
			\includegraphics[width=0.35\linewidth]{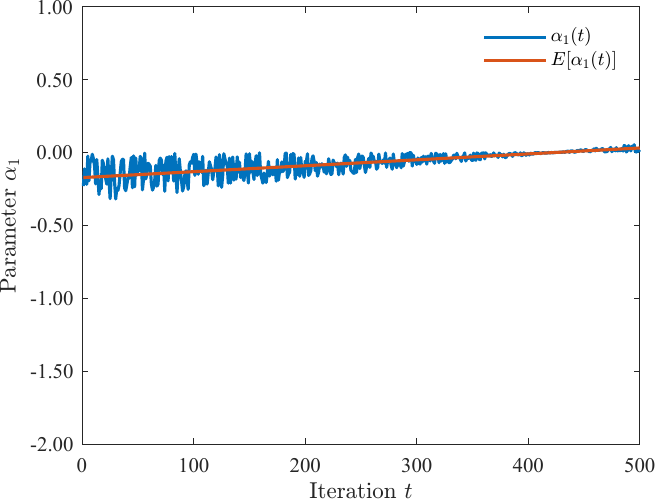}
			\label{fig:alpha_1}}\\
		\subfigure[]{
			\includegraphics[width=0.35\linewidth]{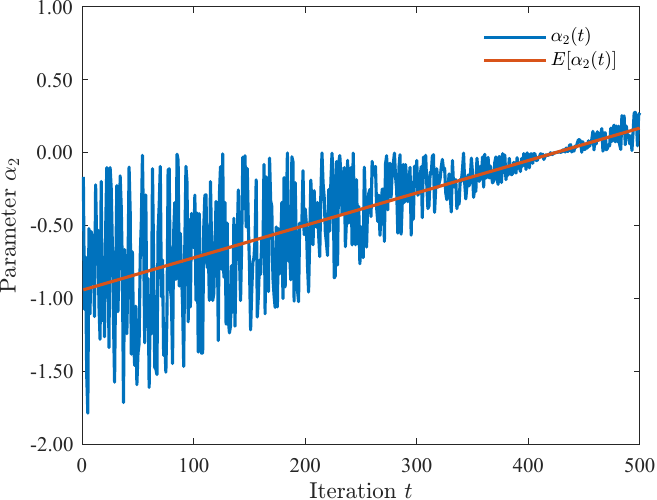}
			\label{fig:alpha_2}}
		\subfigure[]{
			\includegraphics[width=0.35\linewidth]{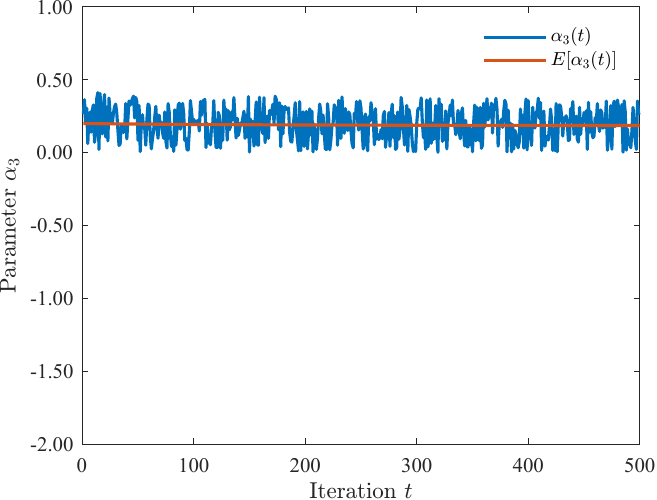}
			\label{fig:alpha_3}}
		\caption{Change of parameters with respect to iteration. (a) $\alpha_1$, (b) $\alpha_2$, (c) $\alpha_3$.}
	\end{figure}

	\paragraph{Sign Mismatch in Parameter $\alpha_1$}
	Equation \eqref{eq:alpha_1} defines the exploration-phase coefficient. The plot of $\alpha_1$ and its expected value with respect to iteration counts is depicted in Fig. \ref{fig:alpha_1}. 
	Given that $d_2 = -d_1$, the ratio $\left| {d_1}/{d_2} \right| = 1$ holds for all $t$, reducing the exponential term to $\exp(-2) = 0.1353$. The expression thus simplifies to:
	\begin{equation}\label{eq:alpha_1_simplify}
		\alpha_1 = 0.4060 r_4 \cdot \left( {t}/{T} - 0.85 \right).
	\end{equation}
	This means that $\alpha_1=0$ when $t/T=0.85$ and there is no update on the agent position. Furthermore, $\alpha_1 < 0$ whenever $t/T < 0.85$, i.e., during the majority of the optimization process. The sign reversal in early iterations contradicts the  narrative that $\alpha_1$ ``steadily stabilizes with decreasing oscillation amplitude.'' Instead, the scaling factor begins negative and grows toward zero, implying early-phase updates may pull away from $\mathbf x_{{best}}$ rather than facilitating convergence. The intended behaviour is thus misaligned with its implementation, introducing potential instability in the initial exploration steps.
	
	\paragraph{Redundant Dynamics in Parameter $\alpha_2$}
	Equation \eqref{eq:alpha_2} defines the coefficient $\alpha_2$. The plot of $\alpha_2$ against iterations is shown in Fig. \ref{fig:alpha_2}. The plot is similar to that of $\alpha_1$ but with a larger magnitude. 
	Since $d_2 = -d_1$, the ratio ${d_1}/{d_2} = -1$ identically for all $t$, and the exponential term simplifies to the constant $\exp(-0.3) = 0.7408$. As a result, $\alpha_2$ reduces to a scaled linear term,
	\begin{equation}\label{eq:alpha__2_simplify}
		\alpha_2 = 2.2225 r_6 \cdot \left( {t}/{T} - 0.85 \right).
	\end{equation}
	This value is negative for most of the optimization process ($t/T < 0.85$), again in tension with the claim that $\alpha_2$ provides smooth convergence behaviour. Moreover, the use of an exponential term and symbolic dependence on $(d_1, d_2)$ gives the illusion of dynamic modulation, when in fact the effect is constant and predetermined. This reflects the same pattern of symbolic over-engineering seen in $\alpha_1$ and $\gamma$, further undermining the transparency and functional purpose of the control coefficients in the algorithm.
	
	\paragraph{Opaque and Ineffective Modulation in Parameter $\alpha_3$}
	The coefficient $\alpha_3$, used in the first exploitation phase, is defined in Eq. \eqref{eq:alpha_3}. Its changes against iteration counts are illustrated in Fig. \ref{fig:alpha_3}.
	By using the expected values at $0.5$ of the random numbers $r_{10}$ and $r_{11}$, and at $t=1$ and $t=T$, we have $\alpha_3$ ranges between $0.1981$ and $0.1846$. This narrow dynamic range renders the modulation effect negligible. The use of nested non-linearities ($\tanh$ inside $\exp$), combined with stochastic masking via $r_{10}$, adds symbolic complexity without functional benefit. As with $\alpha_1$ and $\alpha_2$, the design of $\alpha_3$ exemplifies symbolic over-engineering: a computationally elaborate expression that could be replaced by a simpler, transparent decay schedule without loss of effectiveness or clarity.
	
	\paragraph{Stochastic Saturation and Opaqueness}
	The \ac{ETO} algorithm employs at least 14 independent uniformly distributed random variables per iteration. Many are functionally redundant, reused across multiple update rules without justification, or embedded in expressions whose outcome is structurally indifferent to the randomness (e.g., when multiplied by constant coefficients or static modulating factors). For instance, the random variable $r_6$ governs $\alpha_2$ across both exploration and exploitation phases, yet its repeated use lacks statistical justification or phase-specific decoupling. The cumulative effect is a saturation of stochastic elements that blurs the causal relationship between control logic and observed behaviour.
	
	This overuse of randomness obscures the operative structure, making it difficult to disentangle whether performance gains---if any---stem from the design of the update rules or from random perturbation alone. It also hinders reproducibility: small differences in random sequences can yield divergent outcomes, yet no guidance is provided for seeding strategies or variance control. Furthermore, no empirical ablation study is conducted to evaluate whether such stochastic density is necessary or beneficial. The result is an algorithm whose behavioural trace is noisy by design, but analytically opaque.
	
	\subsection{Improper Agent Update Mechanisms}
	
	The position update rules in \ac{ETO} are partitioned into four phases, each governed by distinct control parameters. We examine their probabilistic behaviour by analysing the resulting probability density functions (\acp{PDF}). The central assumption is that agent positions $x$ and the current best solution $\mathbf x_{\text{best}}$ are uniformly distributed over the search space. Consequently, a properly designed update mechanism should preserve this uniformity, ensuring that no directional bias is introduced and that all regions of the domain remain equally probable for sampling.
	
\begin{figure}[h!]
		\centering
		\subfigure[]{
			\includegraphics[width=0.35\linewidth]{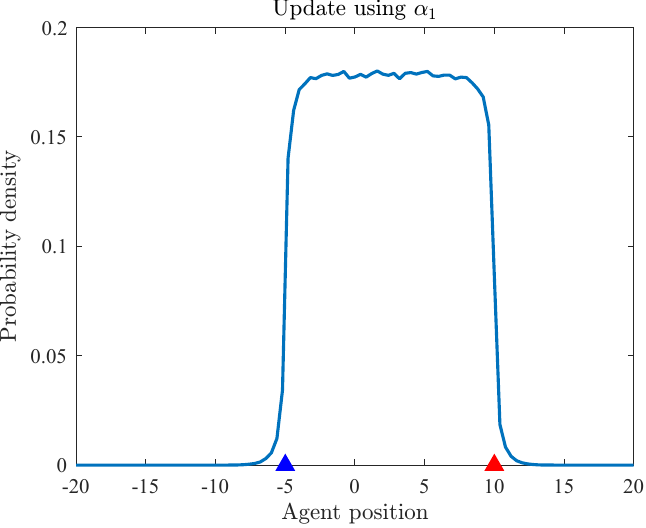}
			\label{fig:Update_a1}}
		\subfigure[]{
			\includegraphics[width=0.35\linewidth]{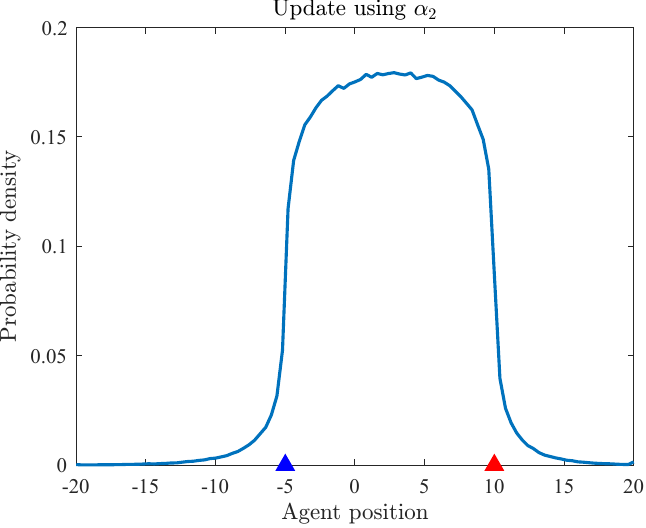}
			\label{fig:Update_a2}}
		\subfigure[]{
			\includegraphics[width=0.35\linewidth]{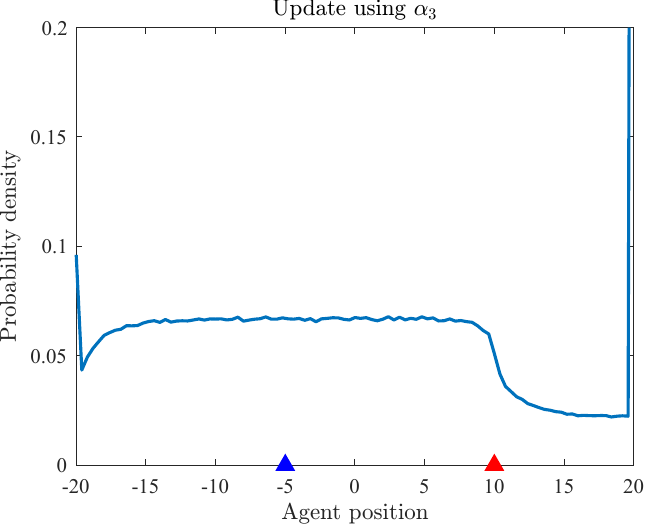}
			\label{fig:Update_a3}}
		\subfigure[]{
			\includegraphics[width=0.35\linewidth]{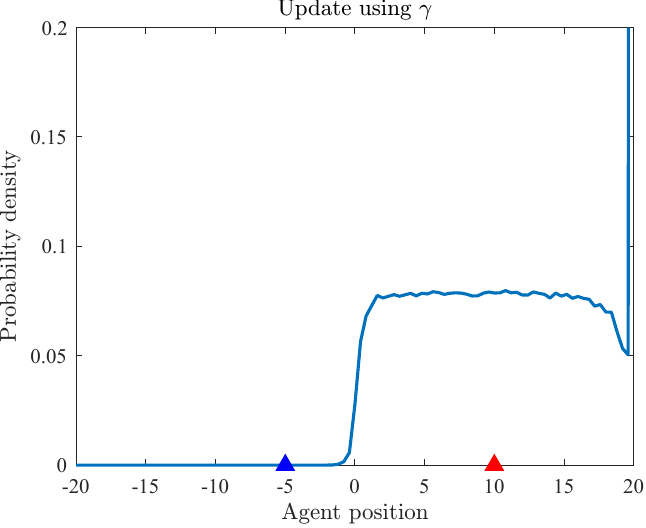}
			\label{fig:Update_gamma}}
		\caption{Empirical \acp{PDF} of agent positions after updates: (a) using $\alpha_1$, (b) using $\alpha_2$, (c) using $\alpha_3$, (d) using $\gamma(t)$. Red and blue markers denote lower and upper bounds, respectively.}
	\end{figure}

	To assess this, $10^6$ samples were generated for each update rule to produce smooth empirical \acp{PDF}. The resolution was set to $0.01$ times the domain length, and the bounds were configured as $L_b = -5$ and $U_b = 10$, introducing intentional asymmetry in the domain. The output distributions from each update mechanism are shown in Fig. \ref{fig:Update_a1}–\ref{fig:Update_gamma}.
	
	Figure \ref{fig:Update_a1} shows the output of the update using $\alpha_1$. The resulting distribution is mildly biased toward the positive half-space, likely due to the asymmetric bounds of the search region. Although visually close to uniform, the distribution mean is shifted away from zero, and a small proportion of agents are pushed beyond the boundaries. Nevertheless, this update generally enables agents to cover the search space with approximately uniform probability.
	
	In contrast, Figure \ref{fig:Update_a2} indicates that the update rule based on $\alpha_2$ results in a marked deterioration from a uniform distribution. The shape more closely resembles a hybrid between a uniform and a truncated normal distribution. This behaviour arises from the use of two random variables, $r_7$ and $r_8$, in a multiplicative form ($r_7 r_8 \alpha_2 x_{\text{best}}$), which compresses the range of variation. As a consequence, the effectiveness of the algorithm in exploring the search space is reduced, and a larger proportion of updated agents fall outside the domain bounds.
	
	Figure \ref{fig:Update_a3} further intensifies this trend. The output distribution is wide and diffuse, exhibiting heavier tails and a significantly higher frequency of boundary violations. According to Eq. \eqref{eq:update_3}, the update magnitude is scaled by a factor of three and involves absolute displacements. This leads to excessive agent movement and increases the likelihood of generating infeasible solutions.
	
	The update governed by $\gamma(t)$, shown in Figure \ref{fig:Update_gamma}, displays pronounced asymmetry. The distribution is almost entirely biased toward the positive region of the domain. This behaviour follows directly from Eq. \eqref{eq:update_4}, where the term $\gamma(t) \cdot |r_{14} x_{\text{best}} - x|$ is strictly non-negative and added to $x$, resulting in systematic rightward shifts. Very few agents remain in the lower half of the search space, which undermines the ability of \ac{ETO} to explore non-positive regions.
	
	In summary, the update mechanisms defined by Eq. \eqref{eq:update_3} and Eq. \eqref{eq:update_4} are particularly problematic. They introduce symbolic bias, directional drift, and a high rate of infeasible outputs. These deficiencies necessitate the use of post-processing mechanisms such as boundary reflection or resampling, yet none are discussed or implemented in the original \ac{ETO} paper. This omission further highlights the lack of implementation rigour and design foresight.
	
	\subsection{Misuse of Descriptive Statistics}
	
	The statistical reporting in the \ac{ETO} paper, like many metaheuristic studies, relies heavily on descriptive summaries, namely the best, mean, and standard deviation of fitness values. This practice represents a systemic issue in metaheuristic literature, where descriptive statistics are used as rhetorical devices rather than scientific evidence.
	
	Reported metrics such as the best, mean, and standard deviation are presented without inferential context (e.g., confidence intervals, hypothesis testing, or stratified variance analysis). While these values summarize performance, they do not quantify uncertainty, significance, or reproducibility. Relying on mean fitness values to claim superiority without statistical testing (e.g., Wilcoxon) renders such claims unfounded, as small differences may arise from stochastic noise or dimensional bias, not genuine algorithmic advantage. Furthermore, highlighting the best-observed fitness across runs exaggerates performance and promotes symbolic inflation by masking instability, especially when the best result is an outlier. Without proper inferential framing, the conclusions remain speculative.

	\section{Comparative Study}\label{sec:comparaive}
	
	This section presents a comparative analysis of the \ac{ETO} algorithm against established metaheuristic baselines. The objective is not merely to report performance metrics, but to diagnose symbolic inflation, benchmarking opacity, and statistical misuse that often accompany such comparisons. All evaluations are grounded in reproducible protocols, with attention to quartile stratification, operator-level attribution, and diagnostic overlays.
	
	\subsection{Experimental Setup}
	
	This study evaluates the performance of ten metaheuristic algorithms across the CEC 2021 and CEC 2017 benchmark suites, encompassing a range of dimensionalities (from 10 to 100 dimensions) and transformation complexities (basic, shifted, and shift-rotated functions). All algorithms were executed over 25 independent runs per function, with 500 iterations and 30 agents per algorithm per run. This configuration ensured consistent runtime behaviour across all benchmark functions and dimensional tiers.
	
	All experiments were conducted using MATLAB R2021b on a Windows 11 (64-bit) operating system. The hardware platform consisted of an Intel(R) Core(TM) i7-8565U CPU @ 1.80 GHz and 8 GB RAM. 
	
	\subsection{Algorithm Selection and Classification}
	
	The ten metaheuristic algorithms evaluated in this study are consistent with those benchmarked in the original \ac{ETO} paper. They were selected to represent a diverse spectrum of algorithmic paradigms, including widely adopted baselines, high-performing heuristics, and recent proposals. 
	
	Two variants of \ac{PSO}, namely \ac{CPS} \cite{bratton2007defining} and \ac{IPS} \cite{shi1998modified}, are specifically chosen to serve as baseline references, enabling comparative diagnosis against both classical and modified agent-based dynamics. Popular algorithms such as \ac{GWO} \cite{mirjalili2014grey} and \ac{SCA} \cite{mirjalili2016sca} are included due to their extensive citation and foundational role in metaheuristic research. Higher-performing algorithms such as \ac{HHO} \cite{heidari2019harris} and \ac{HGS} \cite{yang2021hunger} have demonstrated competitive convergence behaviour and robust performance across multiple benchmarks. Recent algorithms including \ac{AOA} \cite{abualigah2021arithmetic}, \ac{GJO} \cite{chopra2022golden}, and \ac{SCH} \cite{bai2023sinh} reflect emerging trends in nature-inspired or other search strategies and contribute to diversity in operator dynamics.
	
	\subsection{Convergence Behaviours}\label{subsec:convergence}
	
	This section examines the convergence behaviours of the \ac{ETO} algorithm across benchmark functions and dimensional settings. Rather than treating convergence curves as rhetorical artifacts, we interpret them as diagnostic evidence of symbolic inflation, phase transitions, and operator-level dynamics. Annotated plots are used to highlight stagnation zones, cycle-induced perturbations, and the influence of control parameters. All curves are presented with stratified overlays to expose variability and support reproducibility.
	
	\subsubsection{CEC 2021 Functions: Diagnostic Observations}
	
\begin{figure}[!t]
    \centering
    \subfigure[10-dimensional functions.]{
        \begin{minipage}{\linewidth}
            \centering
            \includegraphics[height=0.22\linewidth]{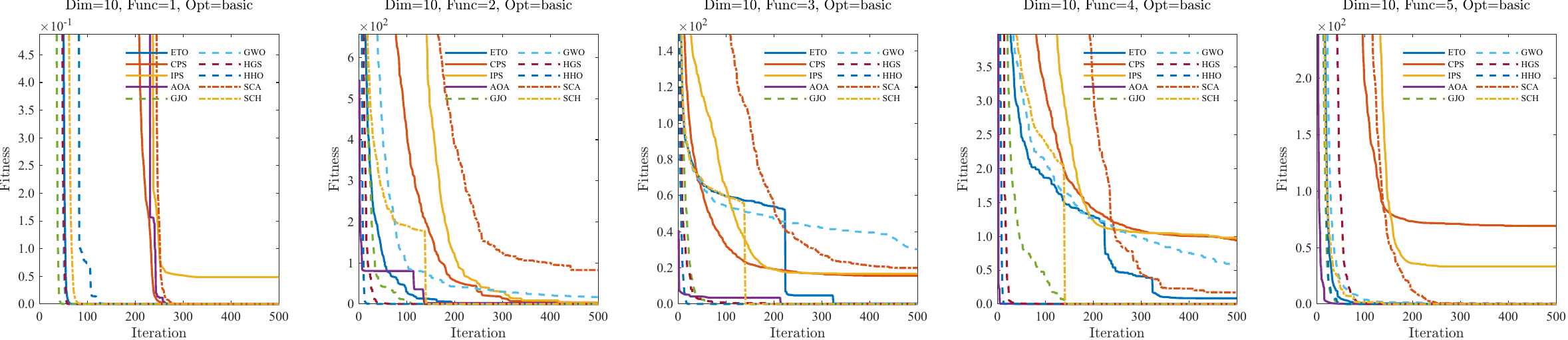}
            \includegraphics[height=0.22\linewidth]{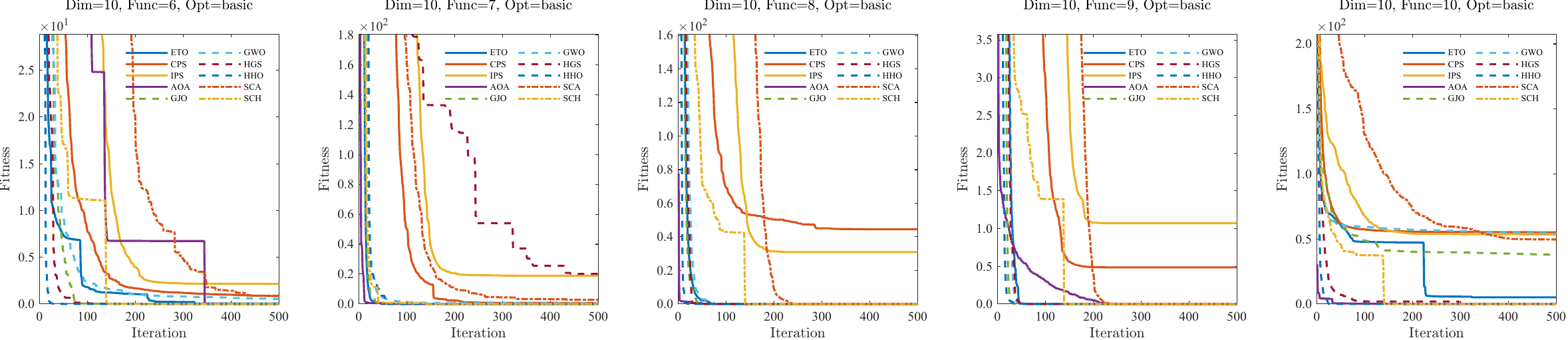}
        \end{minipage}
    }
    \subfigure[20-dimensional functions.]{
        \begin{minipage}{\linewidth}
            \centering
            \includegraphics[height=0.22\linewidth]{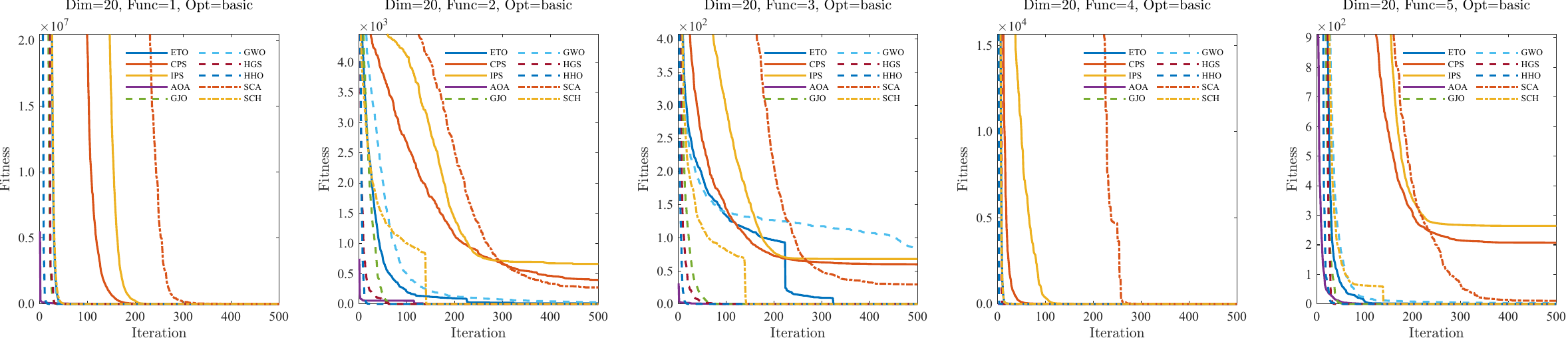}
            \includegraphics[height=0.22\linewidth]{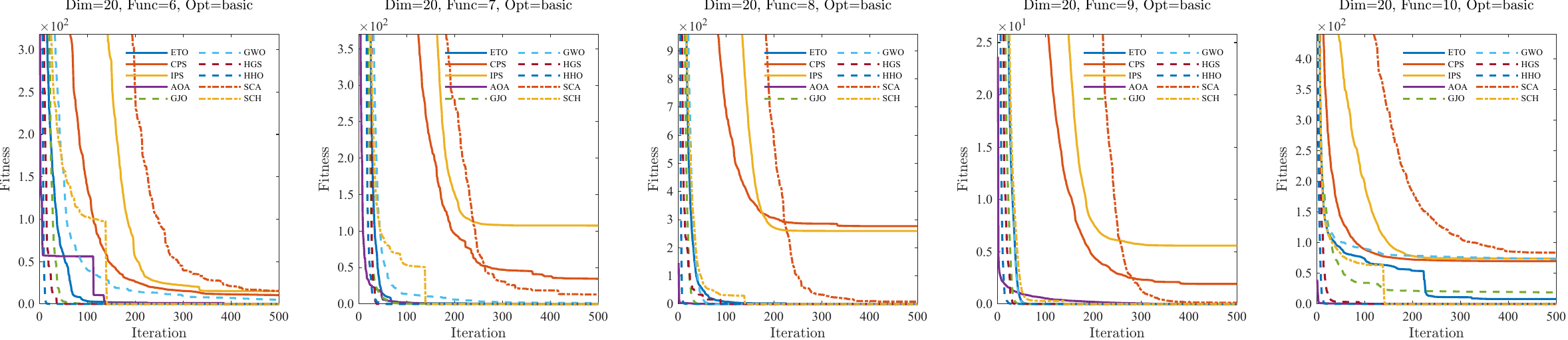}
        \end{minipage}
    }
    \caption{Convergence curves for CEC 2021 basic functions.}
    \label{fig:CEC_2021_Basic}
\end{figure}

\begin{figure}[!t]
    \centering
    \subfigure[10-dimensional functions.]{
        \begin{minipage}{\linewidth}
            \centering
            \includegraphics[height=0.22\linewidth]{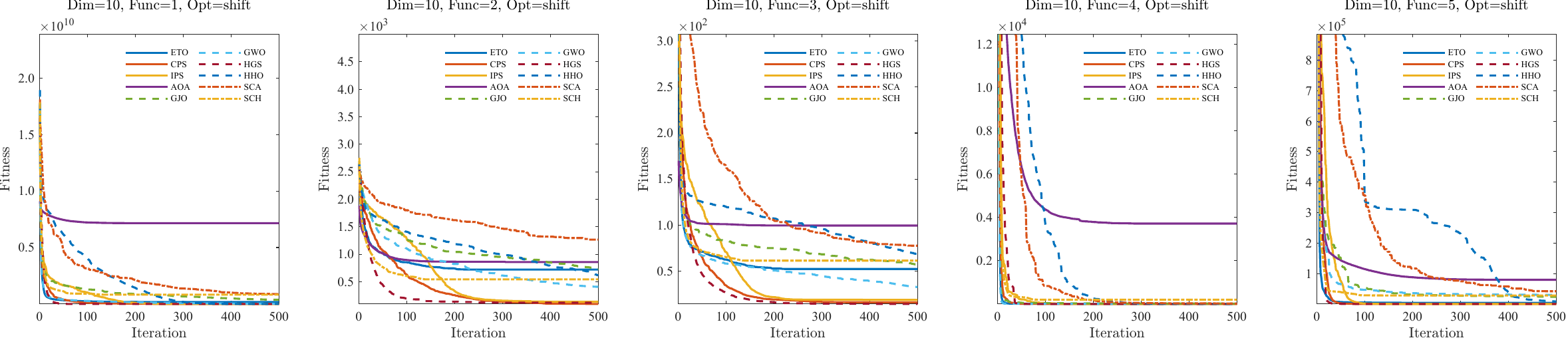}
            \includegraphics[height=0.22\linewidth]{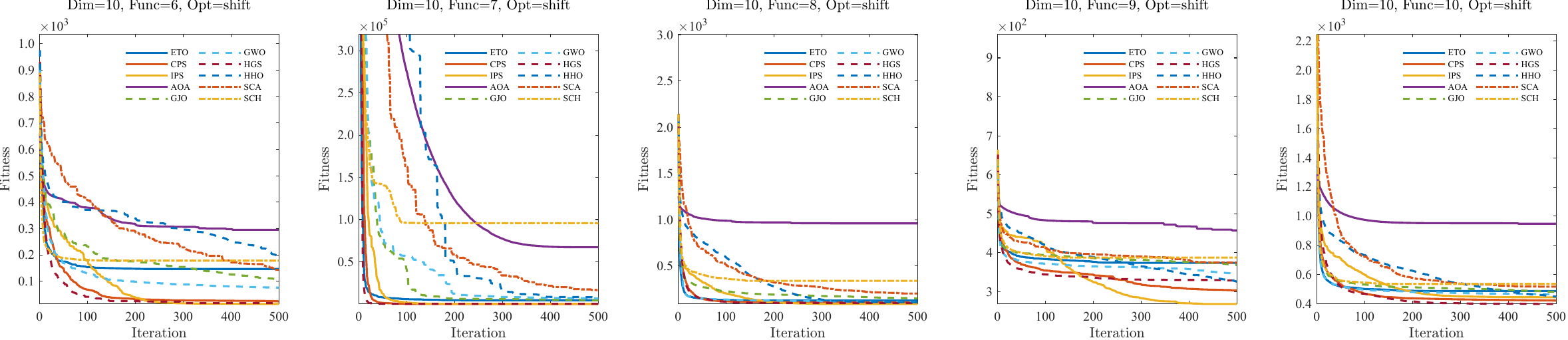}
        \end{minipage}
    }
    \subfigure[20-dimensional functions.]{
        \begin{minipage}{\linewidth}
            \centering
            \includegraphics[height=0.22\linewidth]{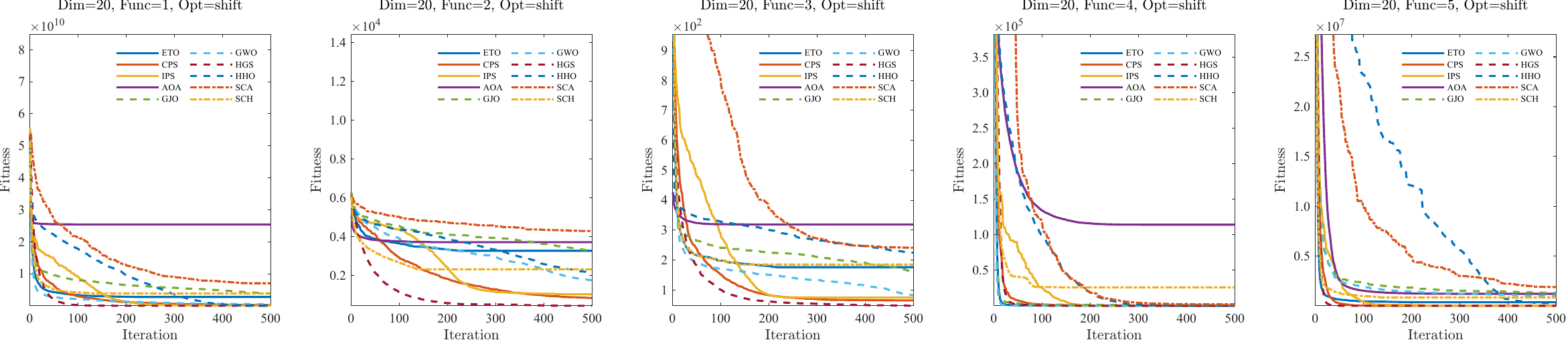}
            \includegraphics[height=0.22\linewidth]{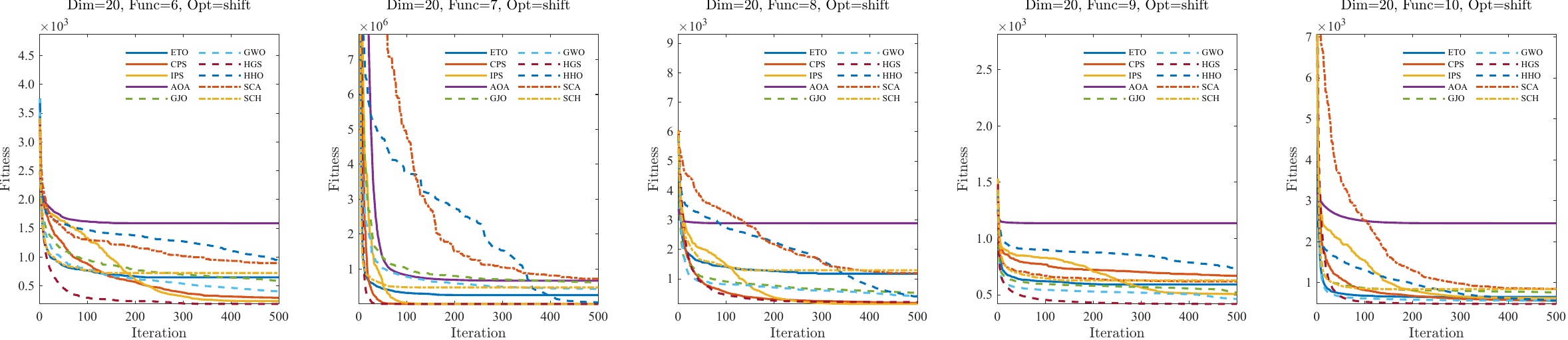}
        \end{minipage}
    }
    \caption{Convergence curves for CEC 2021 shifted functions.}
    \label{fig:CEC_2021_Shift}
\end{figure}

\begin{figure}[!t]
    \centering
    \subfigure[10-dimensional functions.]{
        \begin{minipage}{\linewidth}
            \centering
            \includegraphics[height=0.22\linewidth]{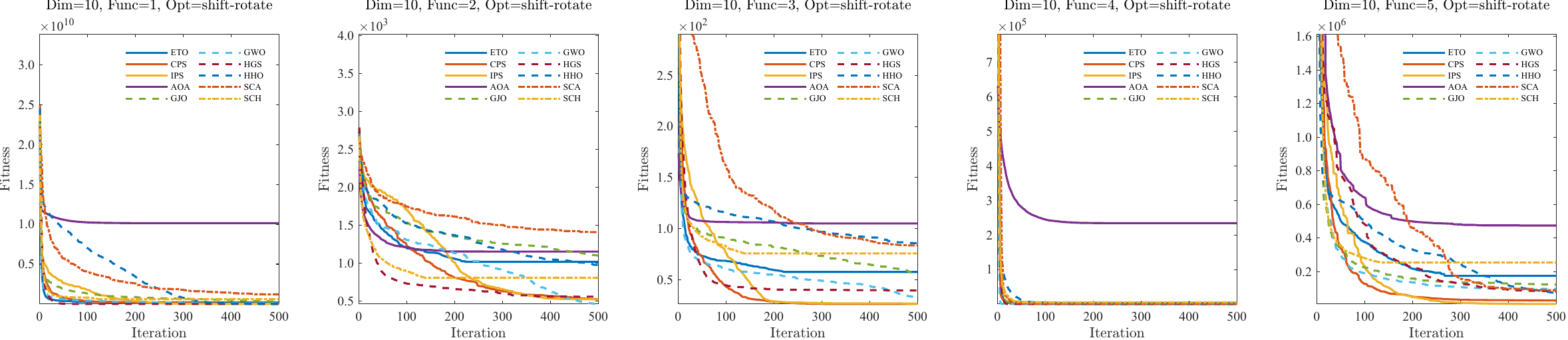}
            \includegraphics[height=0.22\linewidth]{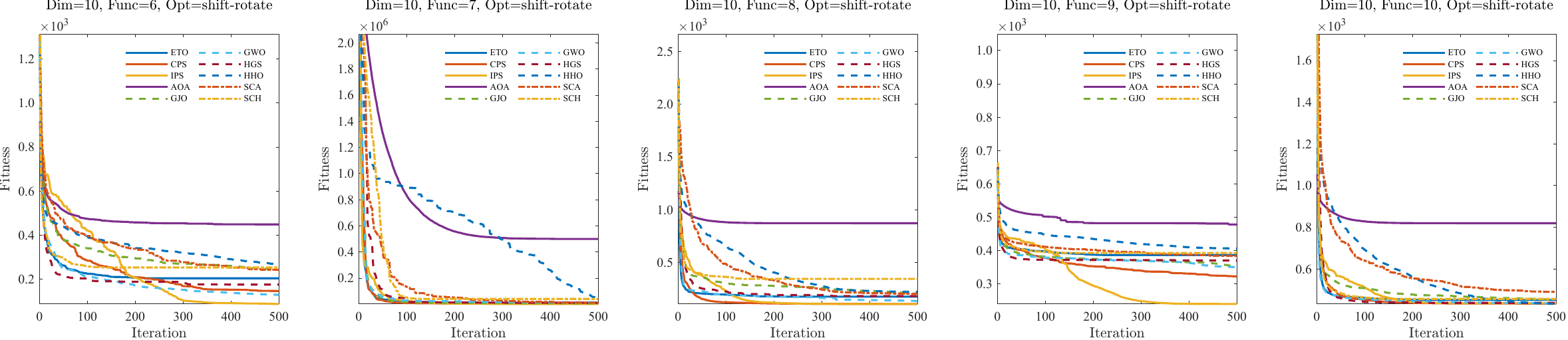}
        \end{minipage}
    }
    \subfigure[20-dimensional functions.]{
        \begin{minipage}{\linewidth}
            \centering
            \includegraphics[height=0.22\linewidth]{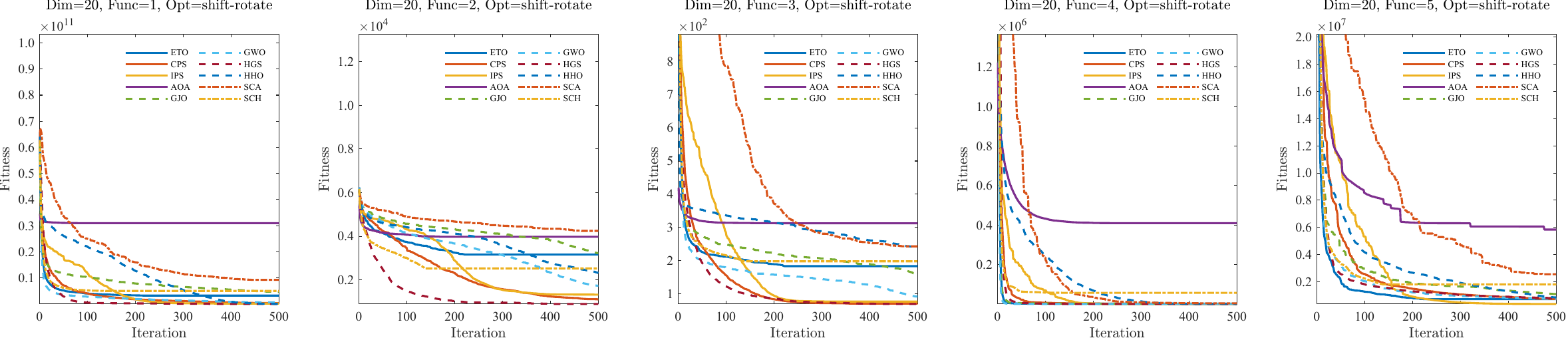}
            \includegraphics[height=0.22\linewidth]{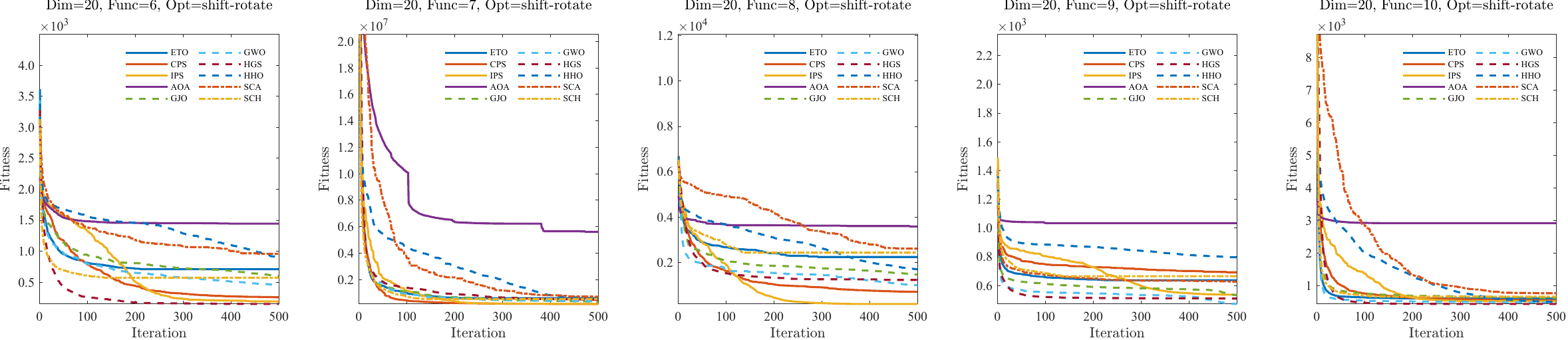}
        \end{minipage}
    }
    \caption{Convergence curves for CEC 2021 shift-rotated functions.}
    \label{fig:CEC_2021_Rotate}
\end{figure}

	Across the CEC 2021 suite (Figs. \ref{fig:CEC_2021_Basic}–\ref{fig:CEC_2021_Rotate}), the \ac{ETO} algorithm demonstrates several consistent diagnostic flaws, all of which are rendered opaque by the lack of stratified evidence (quartile bands, confidence intervals).
	
	\paragraph{Basic Functions}
	The convergence curves in Fig. \ref{fig:CEC_2021_Basic} show steep early descent on unimodal problems (Sphere, Elliptic, Bent Cigar) driven by Phase I operators (Eqs. \ref{eq:alpha_1}–\ref{eq:alpha_3}). This visually impressive descent is diagnostically inflated and typically followed by premature stagnation before the phase transition at $t = \tau$. Non-trivial run-to-run variability on these simple functions indicates instability from the embedded stochastic components ($r_1$–$r_{13}$).
	
	\paragraph{Shifted and Rotated Functions}
	On complex landscapes, Figs. \ref{fig:CEC_2021_Shift} and \ref{fig:CEC_2021_Rotate}, convergence is delayed and often erratic due to poor directional pressure and a lack of rotational invariance. The transition at $t = \tau$ frequently triggers oscillations, flattening, or divergence, suggesting that the modulation coefficient ($\mu$ Eq. \ref{eq:mu_t}) and Phase II updates ($\alpha_2$, Eq. \ref{eq:alpha_2}) lack robustness under translational and rotational biases. Run-to-run overlays show significant divergence, reinforcing the high sensitivity to initialization and the structural fragility of the operators.
	
	\FloatBarrier
	
	\subsubsection{CEC 2017 Functions: Dimensional Instability}
	
\begin{figure}[!t]
    \centering
    \includegraphics[height=0.22\linewidth]{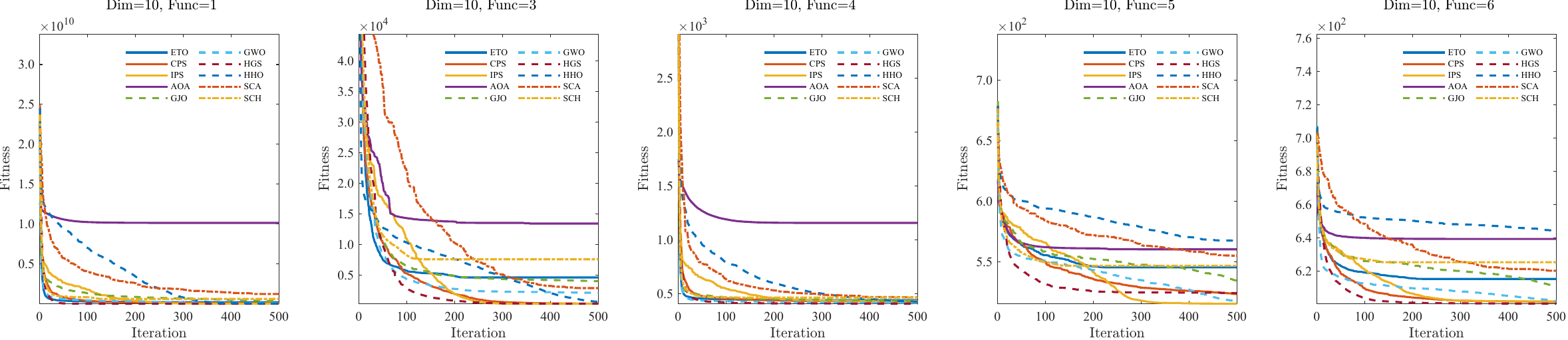}
    \includegraphics[height=0.22\linewidth]{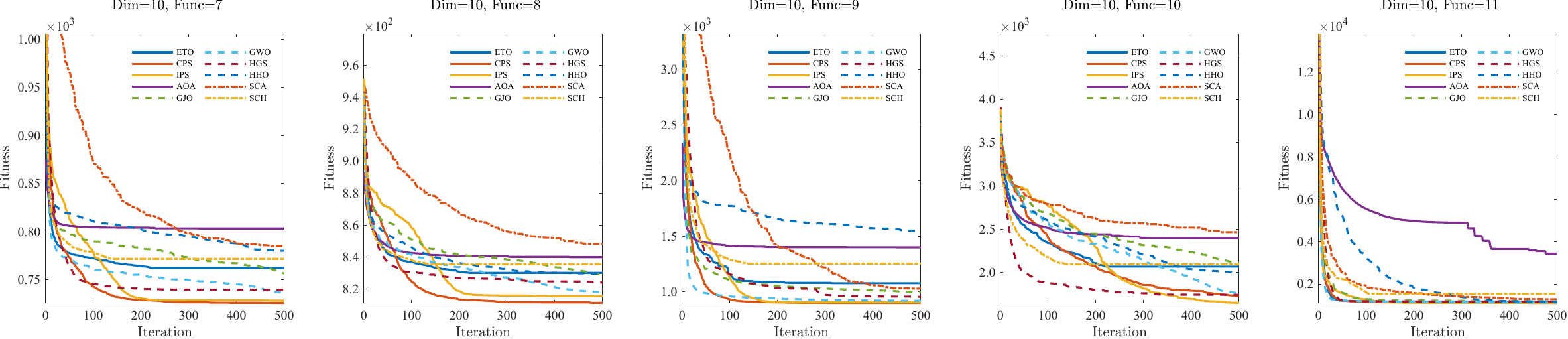}
    \includegraphics[height=0.22\linewidth]{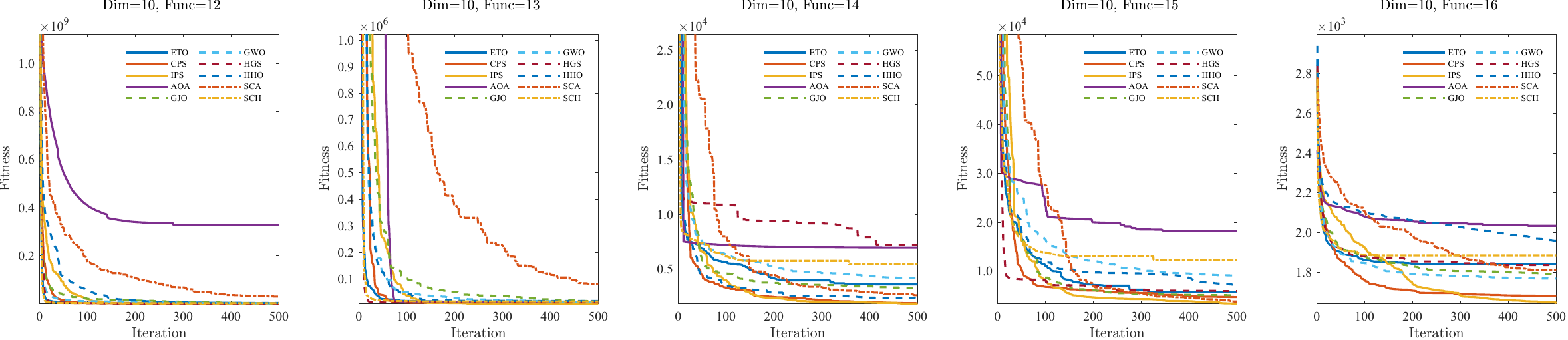}
    \includegraphics[height=0.22\linewidth]{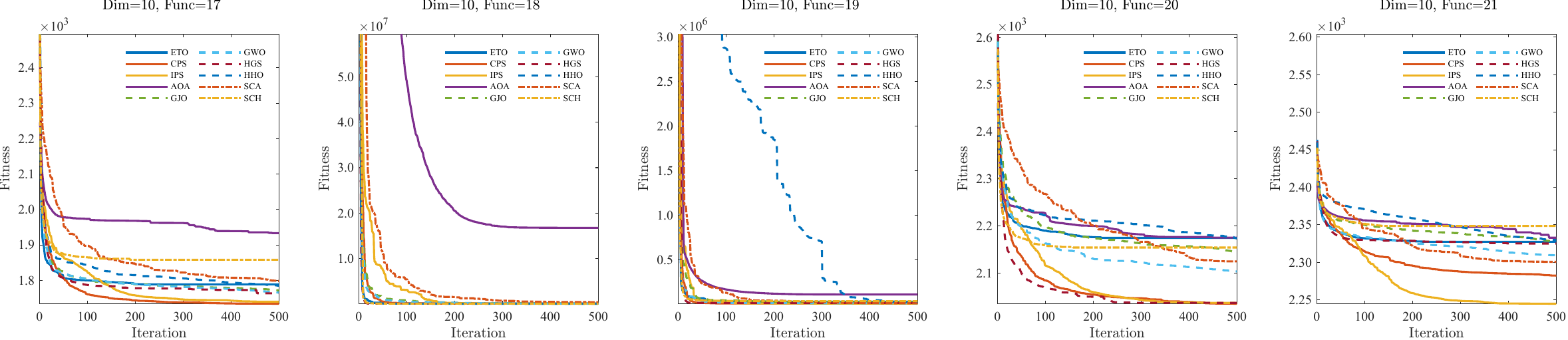}
    \includegraphics[height=0.22\linewidth]{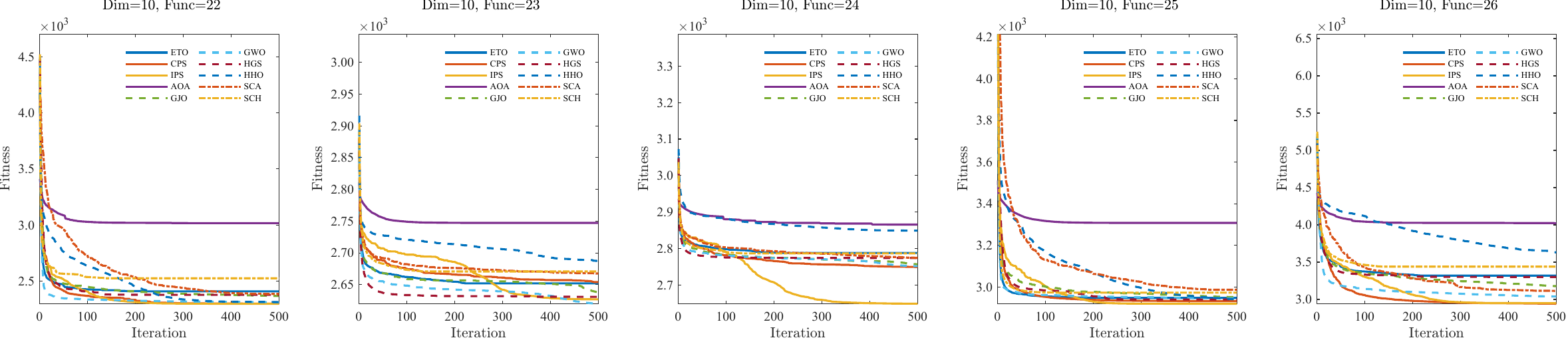}
    \includegraphics[height=0.22\linewidth]{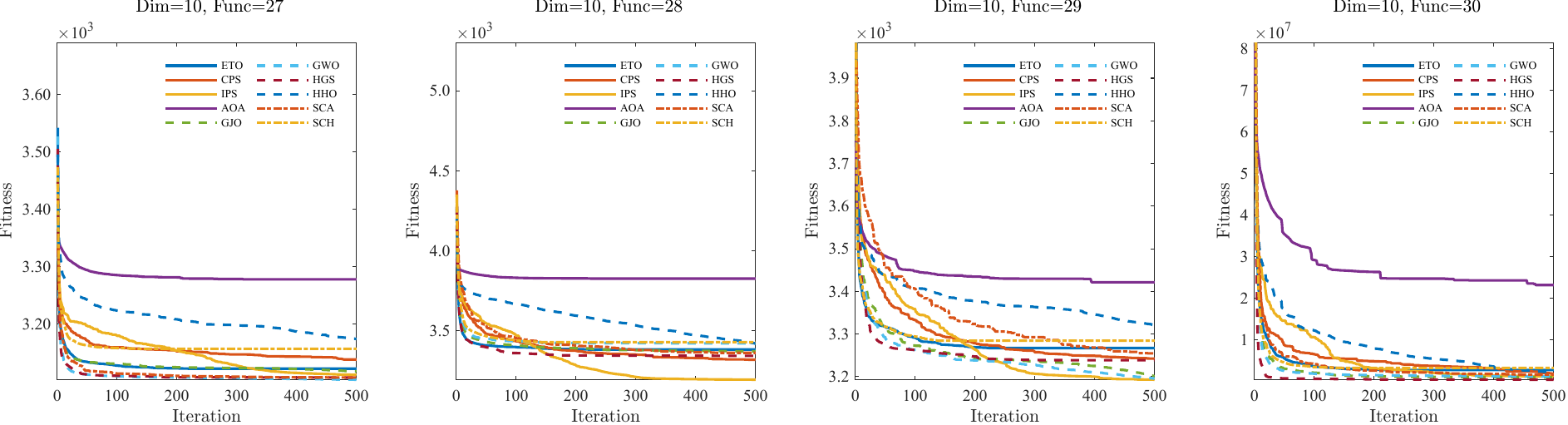}
    \caption{Convergence curves for CEC 2017 10-dimensional functions.}
    \label{fig:CEC_2017_10}
\end{figure}

\begin{figure}[!t]
    \centering
    \includegraphics[height=0.22\linewidth]{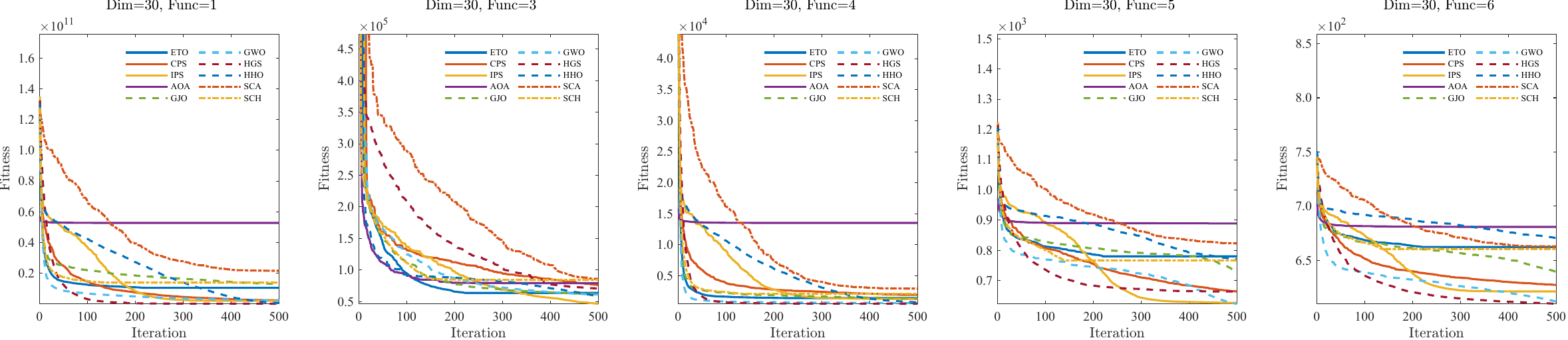}
    \includegraphics[height=0.22\linewidth]{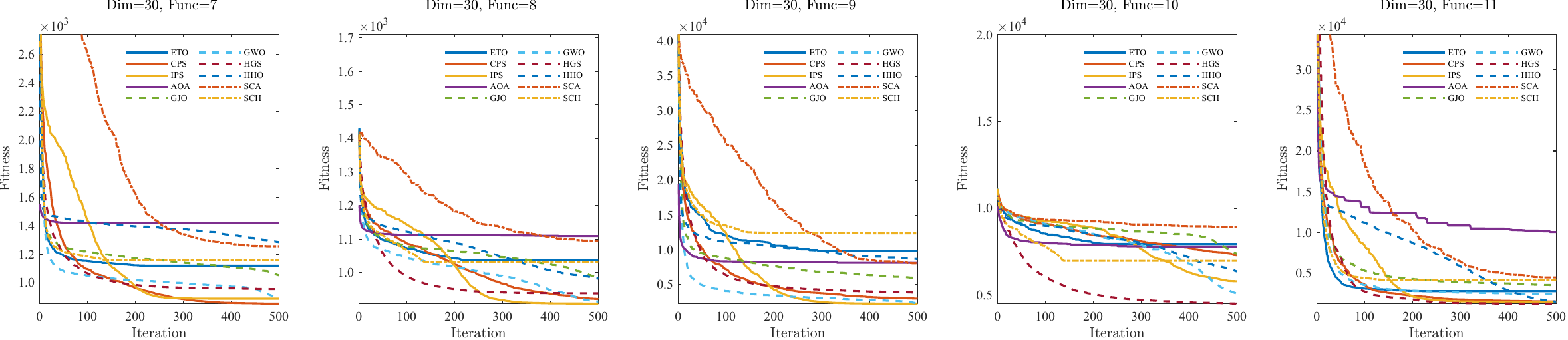}
    \includegraphics[height=0.22\linewidth]{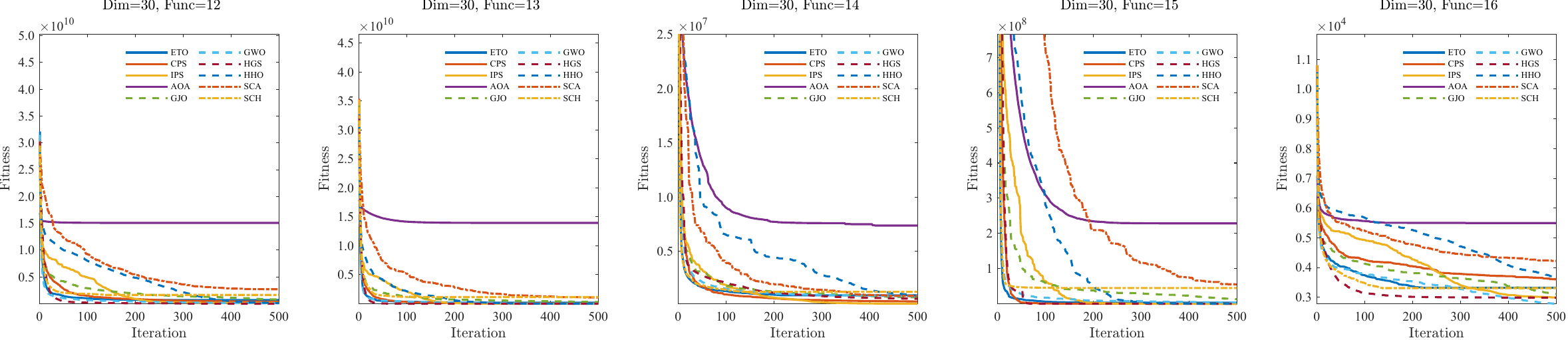}
    \includegraphics[height=0.22\linewidth]{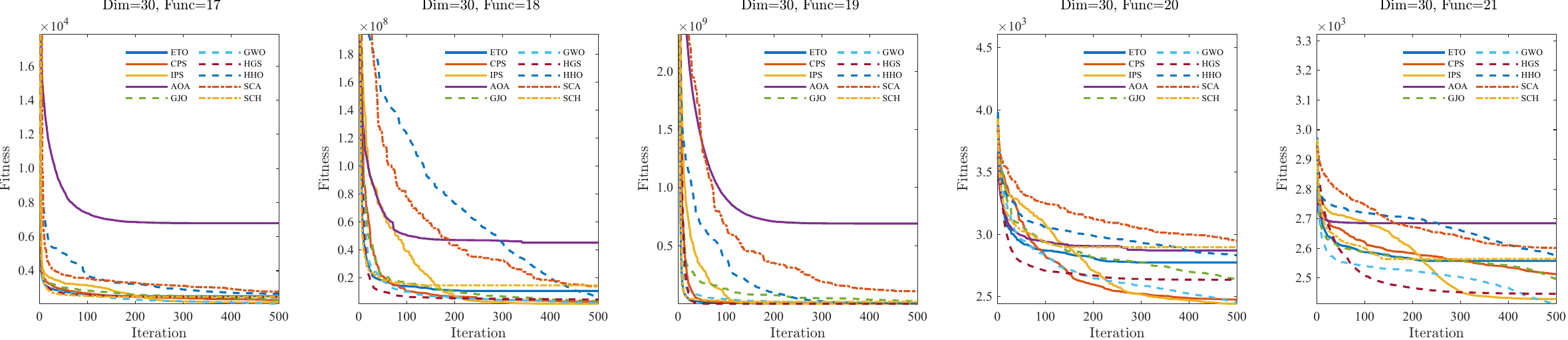}
    \includegraphics[height=0.22\linewidth]{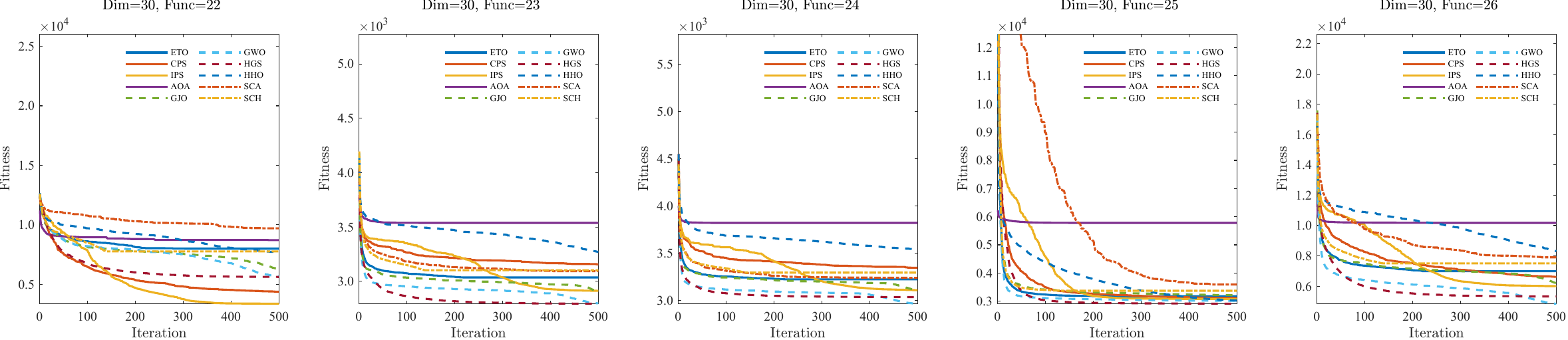}
    \includegraphics[height=0.22\linewidth]{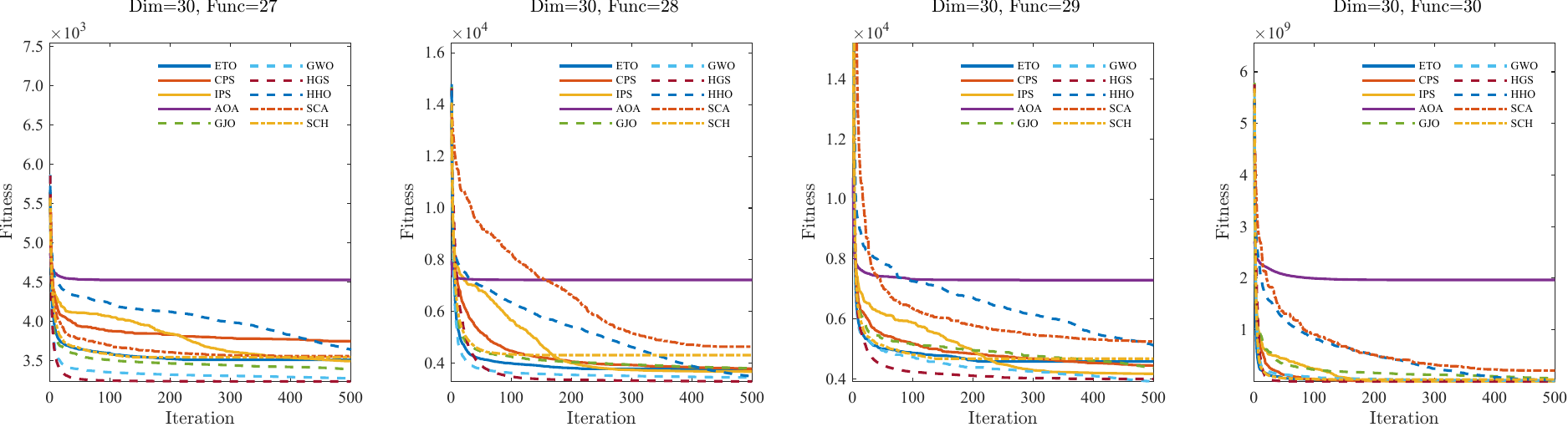}
    \caption{Convergence curves for CEC 2017 30-dimensional functions.}
    \label{fig:CEC_2017_30}
\end{figure}

\begin{figure}[!t]
    \centering
    \includegraphics[height=0.22\linewidth]{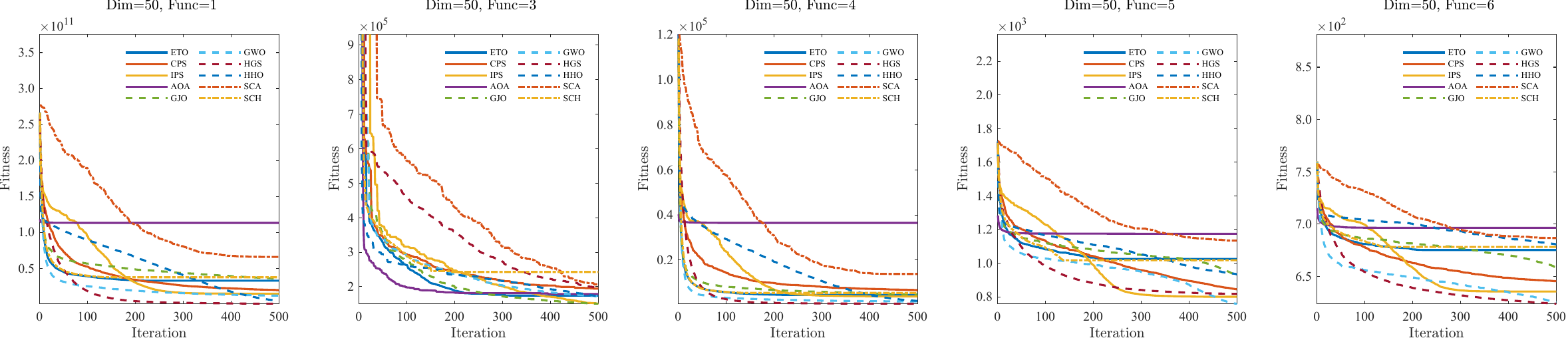}
    \includegraphics[height=0.22\linewidth]{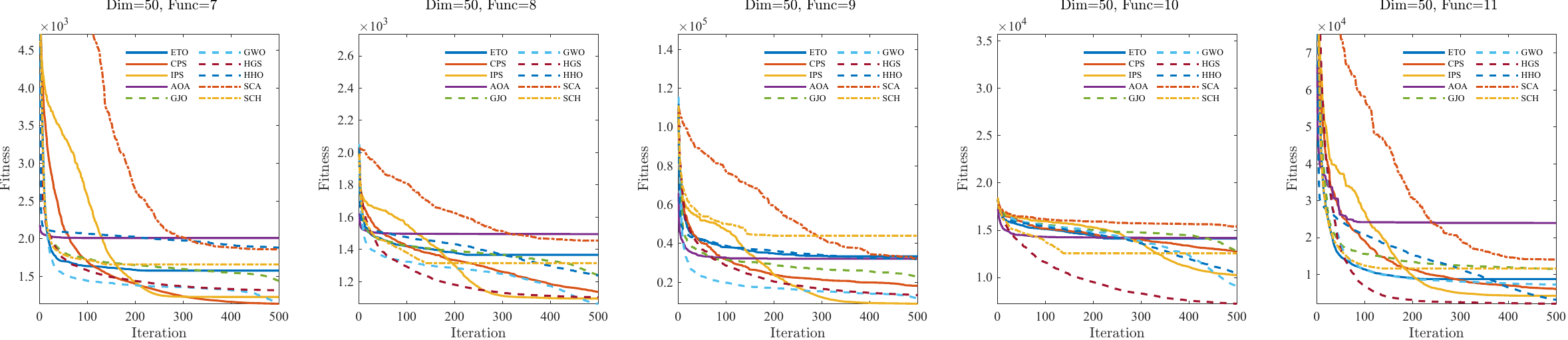}
    \includegraphics[height=0.22\linewidth]{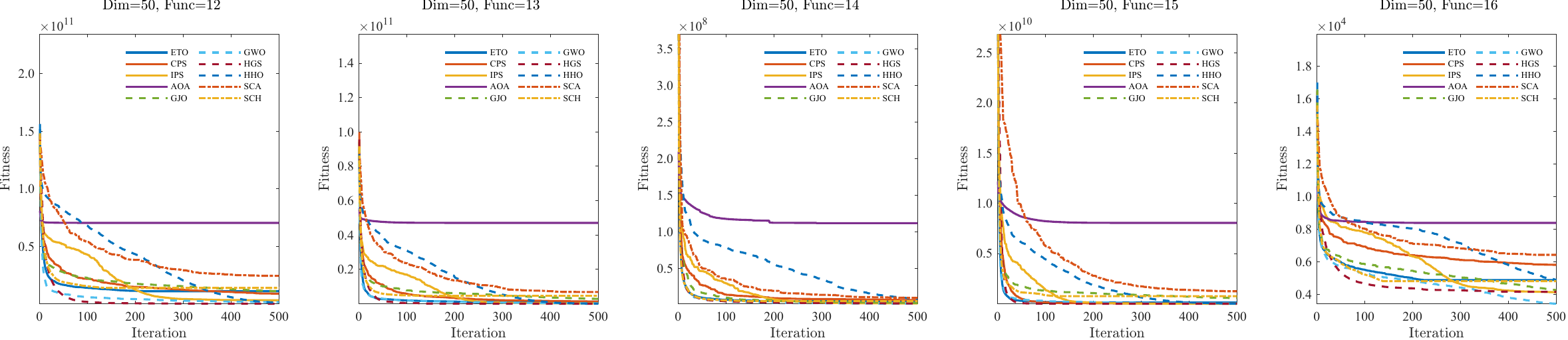}
    \includegraphics[height=0.22\linewidth]{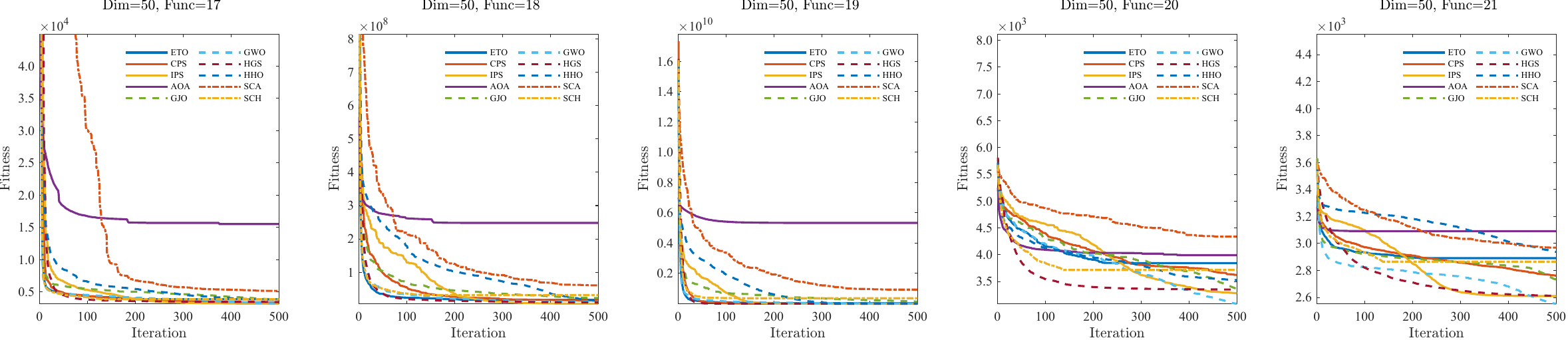}
    \includegraphics[height=0.22\linewidth]{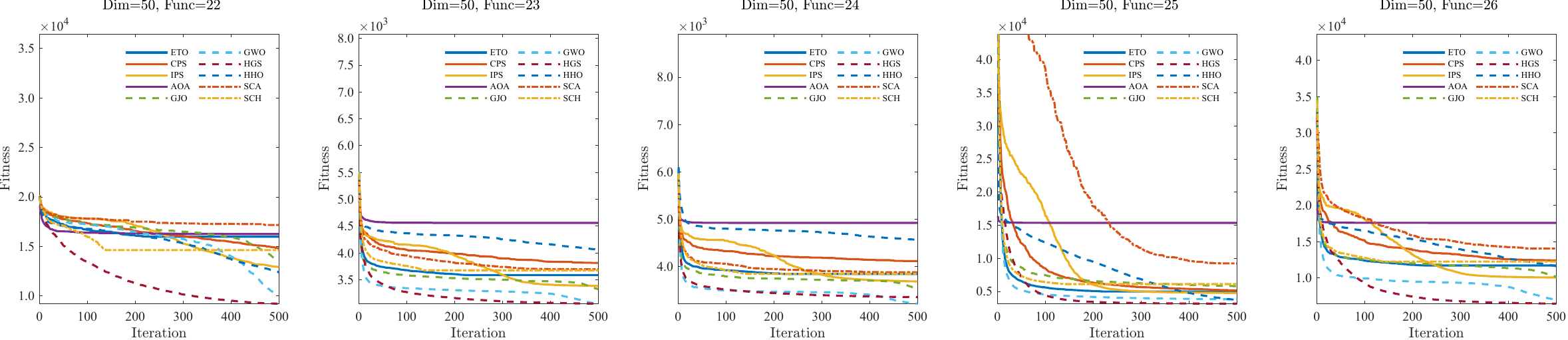}
    \includegraphics[height=0.22\linewidth]{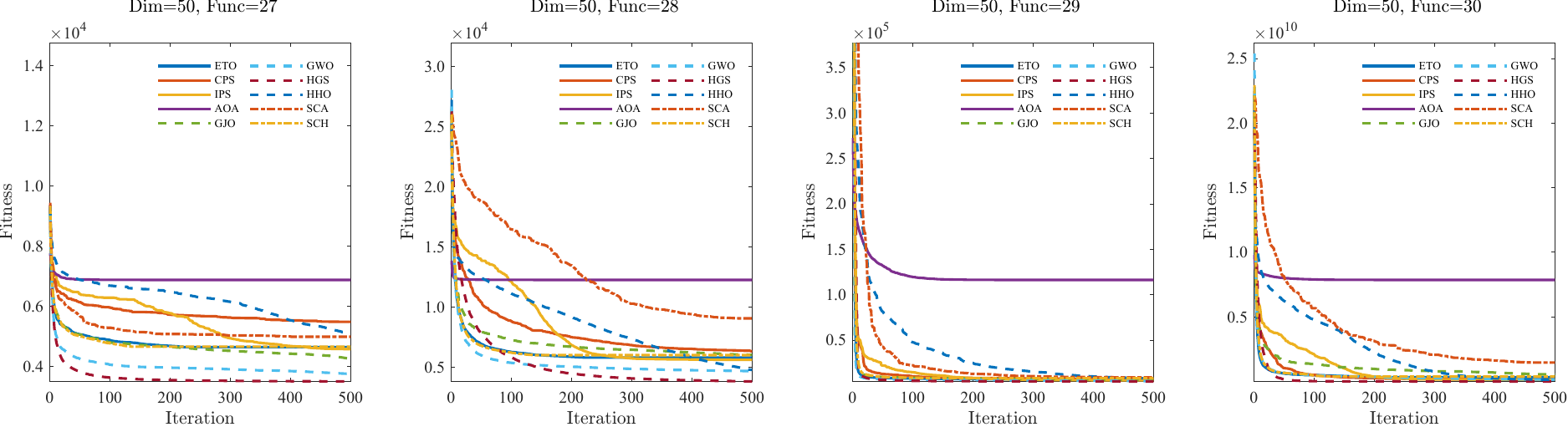}
    \caption{Convergence curves for CEC 2017 50-dimensional functions.}
    \label{fig:CEC_2017_50}
\end{figure}

\begin{figure}[!t]
    \centering
    \includegraphics[height=0.22\linewidth]{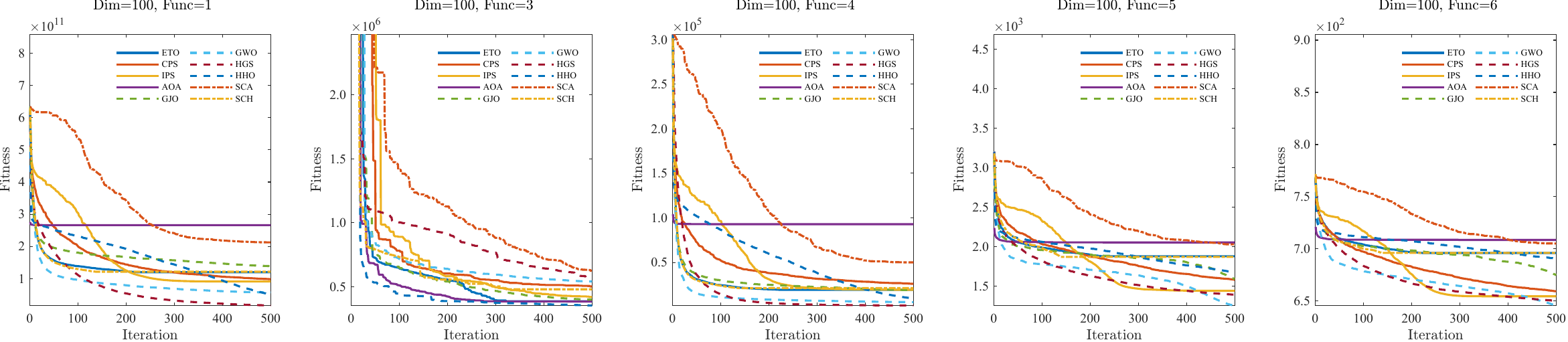}
    \includegraphics[height=0.22\linewidth]{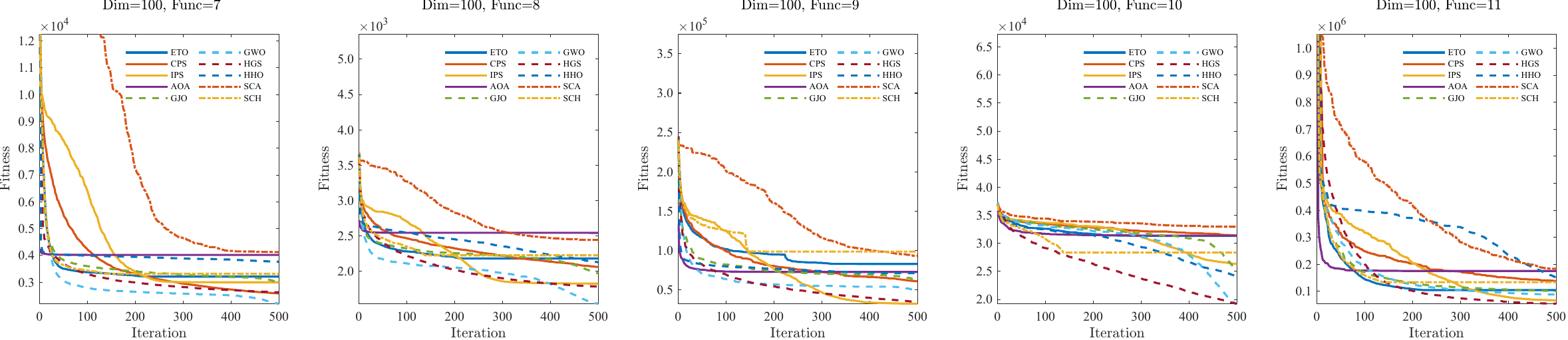}
    \includegraphics[height=0.22\linewidth]{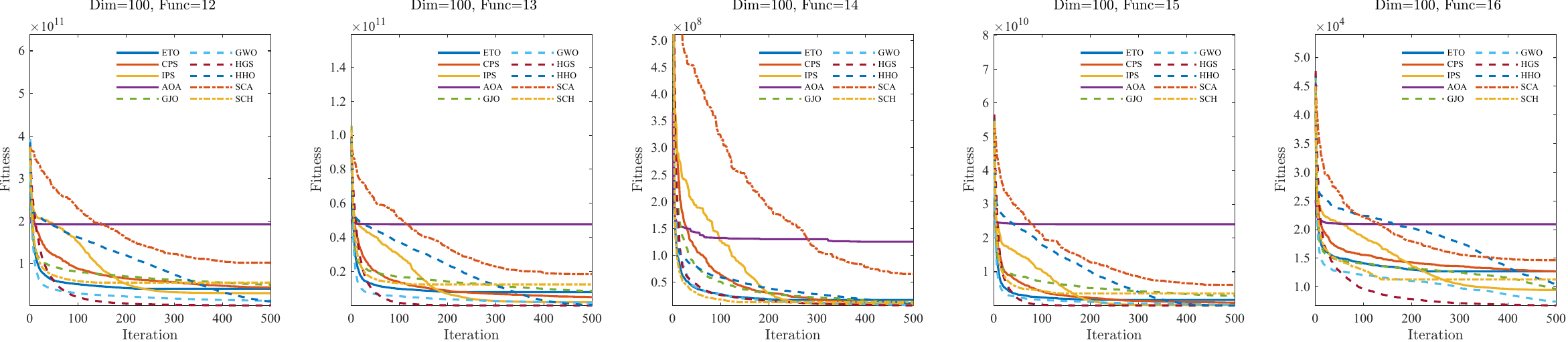}
    \includegraphics[height=0.22\linewidth]{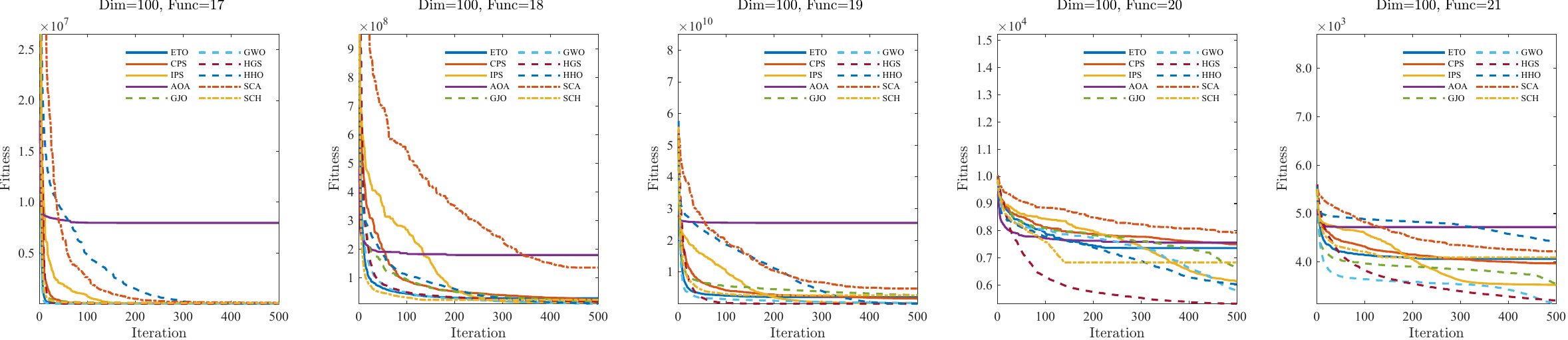}
    \includegraphics[height=0.22\linewidth]{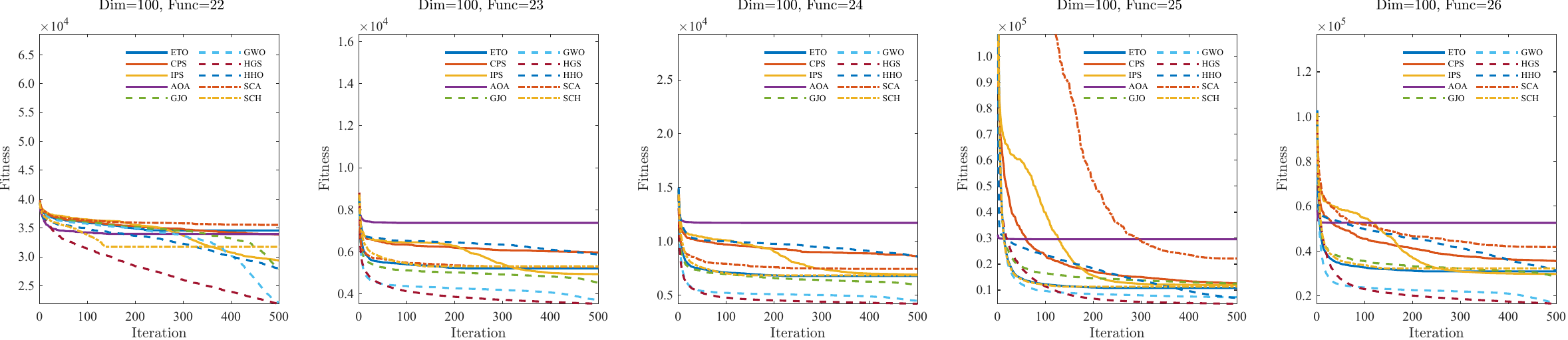}
    \includegraphics[height=0.22\linewidth]{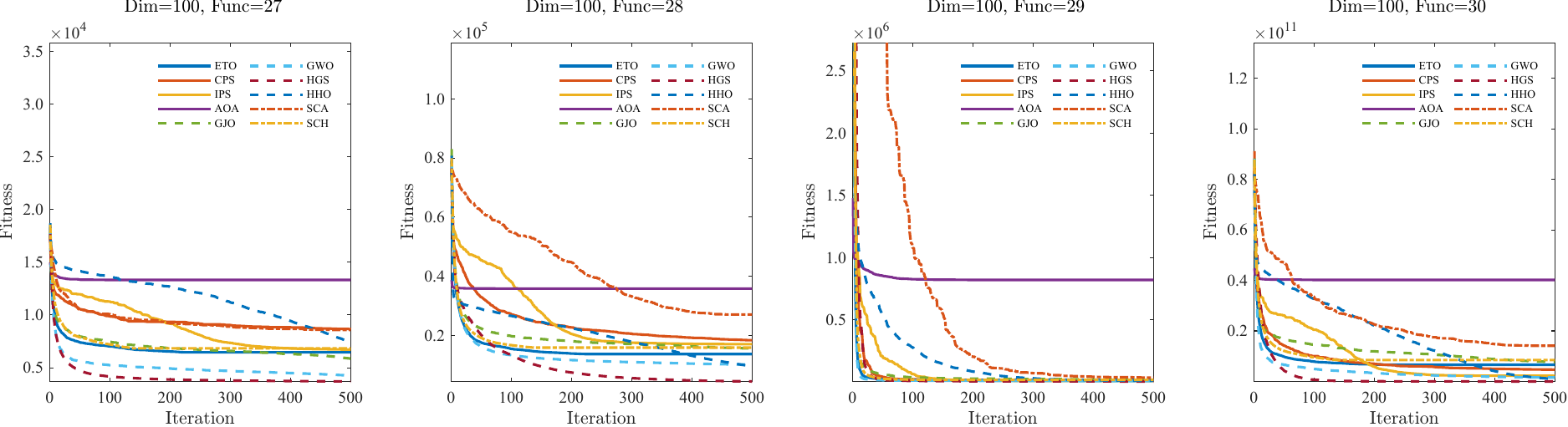}
    \caption{Convergence curves for CEC 2017 100-dimensional functions.}
    \label{fig:CEC_2017_100}
\end{figure}

	Observations across the CEC 2017 functions focus on the algorithm’s scaling behaviour with increasing dimensionality (Figs. \ref{fig:CEC_2017_10}–\ref{fig:CEC_2017_100}).
	
	\paragraph{Low and Mid Dimensions (10- and 30-dimension)}
	In 10 dimensions, Fig. \ref{fig:CEC_2017_10}, Phase I operators induce rapid but inflated early descent, followed by premature stagnation on both unimodal and multimodal functions. At 30 dimensions (Fig. \ref{fig:CEC_2017_30}), symbolic instability becomes pronounced, with curves exhibiting erratic transitions and mid-phase oscillations. The modulation coefficient $\mu$ fails to adapt to the increased complexity, and run-to-run divergence is heightened.
	
	\paragraph{High Dimensions (50- and 100-dimension)}
	The algorithm demonstrates structural fragility in 50 dimensions (Fig. \ref{fig:CEC_2017_50}), with inconsistent early descent and fragmented convergence on composition functions. At 100 dimensions (Fig. \ref{fig:CEC_2017_100}), the algorithm's limitations are fully exposed: curves stagnate early with minimal improvement beyond initial iterations. Excessive randomness ($r_1$–$r_{13}$) and symbolic over-design interact destructively, resulting in noisy, unstable trajectories that fail to demonstrate scalability or reproducibility.
	
	\subsection{Summary of Diagnostic Findings}
	
	Across all benchmark categories and dimensional settings, the convergence curves of the \ac{ETO} algorithm exhibit three core methodological flaws:
	\begin{enumerate}
		\item Symbolic Inflation and Stagnation: Early-phase descent is artificially inflated by non-linear operators (Eqs. \ref{eq:alpha_1}–\ref{eq:alpha_3}), yielding visually impressive but methodologically weak trajectories that consistently collapse into mid-phase stagnation near $t=\tau$.
		\item Structural Fragility: The algorithm displays poor robustness under landscape complexity (shifted/rotated functions) and high dimensionality, where the modulation coefficient $\mu$ and Phase II operators ($\alpha_2$) interact unpredictably, leading to high run-to-run variability and erratic convergence paths.
		\item Diagnostic Opacity: The visual evidence is anecdotal and prone to over-interpretation due to the consistent absence of stratified diagnostics (quartile bands, confidence intervals) and operator-level attribution (ablation markers). This opacity allows the curves to function as rhetorical embellishment rather than scientific validation, reinforcing the need for symbolic hygiene in reporting.
	\end{enumerate}
	
	\FloatBarrier
	
	\FloatBarrier
	
	\subsection{Statistical Procedures and Metrics}\label{subsec:statistical}
	
	To quantify algorithmic differences, the Friedman test was employed to detect global rank variance, followed by pairwise comparisons using the Dunn-Sidak corrected Wilcoxon signed-rank test. Effect size metrics, including Cliff's $\delta$ and standardized rank correlation, were reported to characterize the magnitude and direction of performance gaps. Additionally, quartile stratification was applied to average ranks to facilitate symbolic tagging and diagnostic overlays. Median differences were included to support convergence curve interpretation and highlight structural disparities.
	
	This framework integrates rank-based statistics, effect size diagnostics, and quartile tagging, enabling principled benchmarking, symbolic hygiene, and transparent operator profiling across diverse optimization landscapes. These rigorous procedures support reproducible attribution and editorial clarity, thereby aligning with reformist goals in metaheuristic evaluation.
	
	\subsection{CEC 2021 Functions}
	
	\subsubsection{Basic Functions}
	
\begin{table}[!htbp]
\centering
\caption{Results of statistical test for CEC 2021 10-dimensional basic functions.}
\label{tab:Table_CEC_2021_10_Basic}
\footnotesize

\begin{tabular}{lrl}
\toprule
\multicolumn{3}{c}{Friedman test} \\
\midrule
Statistic & Value & Interpretation \\
\midrule
$\chi^2(9, N=250)$ & 1760.808 & Test statistic for rank variance \\
$p$-value & 0.000 & Strong evidence against $H_0$ \\
Kendall's $W$ & 0.783 & High concordance among rankings \\
\end{tabular}

\begin{tabular}{lrrrrr}
\toprule
\multicolumn{6}{c}{Average Ranks and Quartiles} \\
\midrule
& ETO & CPS & IPS & AOA & GJO \\
\midrule
Average Rank & 3.460 & 8.720 & 8.936 & 2.896 & 5.028 \\
Quartile     & 2     & 4     & 4     & 1     & 3 \\
\midrule
& GWO & HGS & HHO & SCA & SCH \\
\midrule
Average Rank & 7.344 & 3.684 & 4.164 & 8.156 & 2.612 \\
Quartile     & 3     & 2     & 2     & 4     & 1 \\
\end{tabular}

\begin{tabular}{llrrrrr}
\toprule
\multicolumn{7}{c}{Post hoc pair-wise comparison (Dunn-Sidak corrected)} \\
\midrule
Group 1 & Group 2 & Adjusted P & Wilcoxon P & Effect size $r$ & Cliff's $\delta$ & Median Diff. \\
\midrule
ETO & CPS & 0.000 & 0.000 & 0.867 & $-1.000$ & $-1.152$ \\
ETO & IPS & 0.000 & 0.000 & 0.859 & $-0.976$ & $-2.086$ \\
ETO & AOA & 0.707 & 0.000 & 0.578 & 0.190 & 0.000 \\
ETO & GJO & 0.000 & 0.001 & 0.258 & $-0.533$ & $-0.000$ \\
ETO & GWO & 0.000 & 0.000 & 0.785 & $-0.909$ & $-0.014$ \\
ETO & HGS & 1.000 & 0.069 & 0.172 & $-0.196$ & 0.000 \\
ETO & HHO & 0.229 & 0.826 & 0.017 & $-0.420$ & 0.000 \\
ETO & SCA & 0.000 & 0.000 & 0.674 & $-0.839$ & $-0.000$ \\
ETO & SCH & 0.039 & 0.000 & 0.871 & 1.000 & 0.000 \\
\bottomrule
\end{tabular}

\normalsize
\end{table}

\begin{table}[!htbp]
\centering
\caption{Results of statistical test for CEC 2021 20-dimensional basic functions.}
\label{tab:Table_CEC_2021_20_Basic}
\footnotesize

\begin{tabular}{lrl}
\toprule
\multicolumn{3}{c}{Friedman test} \\
\midrule
Statistic & Value & Interpretation \\
\midrule
$\chi^2(9, N=250)$ & 1878.093 & Test statistic for rank variance \\
$p$-value & 0.000 & Strong evidence against $H_0$ \\
Kendall's $W$ & 0.835 & High concordance among rankings \\
\end{tabular}

\begin{tabular}{lrrrrr}
\toprule
\multicolumn{6}{c}{Average Ranks and Quartiles} \\
\midrule
& ETO & CPS & IPS & AOA & GJO \\
\midrule
Average Rank & 3.290 & 8.624 & 9.296 & 3.266 & 5.028 \\
Quartile     & 2     & 4     & 4     & 2     & 3 \\
\midrule
& GWO & HGS & HHO & SCA & SCH \\
\midrule
Average Rank & 7.154 & 3.046 & 4.210 & 8.436 & 2.650 \\
Quartile     & 3     & 1     & 2     & 4     & 1 \\
\end{tabular}

\begin{tabular}{llrrrrr}
\toprule
\multicolumn{7}{c}{Post hoc pair-wise comparison (Dunn-Sidak corrected)} \\
\midrule
Group 1 & Group 2 & Adjusted P & Wilcoxon P & Effect size $r$ & Cliff's $\delta$ & Median Diff. \\
\midrule
ETO & CPS & 0.000 & 0.000 & 0.865 & $-0.992$ & $-12.129$ \\
ETO & IPS & 0.000 & 0.000 & 0.867 & $-1.000$ & $-60.982$ \\
ETO & AOA & 1.000 & 0.161 & 0.135 & $-0.296$ & $0.000$ \\
ETO & GJO & 0.000 & 0.000 & 0.281 & $-0.581$ & $-0.000$ \\
ETO & GWO & 0.000 & 0.000 & 0.790 & $-0.918$ & $-0.093$ \\
ETO & HGS & 1.000 & 0.000 & 0.381 & 0.046 & $0.000$ \\
ETO & HHO & 0.014 & 0.080 & 0.137 & $-0.509$ & $0.000$ \\
ETO & SCA & 0.000 & 0.000 & 0.817 & $-0.944$ & $-2.937$ \\
ETO & SCH & 0.428 & 0.000 & 0.872 & 1.000 & $0.000$ \\
\bottomrule
\end{tabular}

\normalsize
\end{table}

	For the 10-dimensional basic functions, Table \ref{tab:Table_CEC_2021_10_Basic}, the Friedman test yielded $\chi^2(9, N=250) = 1760.808$ ($p = 0.000$) with Kendall's $W = 0.783$, indicating high concordance among algorithmic rankings. \ac{HGS} and \ac{SCH} emerged as top performers, dominating the first quartile (average ranks 3.684 and 2.612). \ac{ETO} ranked third overall (3.460) and fell into the second quartile, outperforming CPS, IPS, and \ac{GJO}. Pairwise comparisons showed statistically significant advantages of \ac{ETO} over CPS and IPS ($p < 0.001$), supported by large effect sizes ($r > 0.85$) and Cliff's $\delta$ approaching $-1.000$. However, \ac{ETO}'s performance was statistically indistinguishable from \ac{AOA} and \ac{HGS}, suggesting convergence in solution quality among mid-ranked operators.
	
	In the 20-dimensional case, Table \ref{tab:Table_CEC_2021_20_Basic}, the Friedman test reported $\chi^2(9, N=250) = 1878.093$ ($p = 0.000$), reinforcing strong rank consistency ($W = 0.835$). \ac{HGS} and \ac{SCH} again dominated the first quartile, while \ac{ETO} maintained a competitive position with an average rank of 3.290 (second quartile). Notably, \ac{ETO} significantly outperformed CPS, IPS, and \ac{GWO} with large effect sizes ($r > 0.79$) and median differences exceeding 60.0 units. Its performance against \ac{AOA} was statistically neutral ($p = 1.000$), while comparisons with \ac{HGS} and \ac{HHO} yielded mixed results, reflecting nuanced operator-level interactions.
	
	Overall, the CEC 2021 basic function results highlight that \ac{ETO} demonstrated reliable mid-tier performance with statistically significant wins over several fourth-quartile heuristics, yet lacked dominance over top-tier operators like \ac{HGS} and \ac{SCH}. These outcomes underscore the importance of phased operator profiling and convergence curve overlays to triangulate algorithmic behaviour across dimensional scales.
	
	\subsubsection{Shifted Functions}
	
\begin{table}[!htbp]
\centering
\caption{Results of statistical test for CEC 2021 10-dimensional shifted functions.}
\label{tab:Table_CEC_2021_10_Shifted}
\footnotesize

\begin{tabular}{lrl}
\toprule
\multicolumn{3}{c}{Friedman test} \\
\midrule
Statistic & Value & Interpretation \\
\midrule
$\chi^2(9, N=250)$ & 1537.800 & Test statistic for rank variance \\
$p$-value & 0.000 & Strong evidence against $H_0$ \\
Kendall's $W$ & 0.683 & Moderate-to-high concordance \\
\end{tabular}

\begin{tabular}{lrrrrr}
\toprule
\multicolumn{6}{c}{Average Ranks and Quartiles} \\
\midrule
& ETO & CPS & IPS & AOA & GJO \\
\midrule
Average Rank & 6.336 & 2.668 & 2.388 & 9.280 & 6.352 \\
Quartile     & 3     & 2     & 1     & 4     & 3 \\
\midrule
& GWO & HGS & HHO & SCA & SCH \\
\midrule
Average Rank & 4.776 & 1.944 & 6.164 & 8.248 & 6.844 \\
Quartile     & 2     & 1     & 2     & 4     & 4 \\
\end{tabular}

\begin{tabular}{llrrrrr}
\toprule
\multicolumn{7}{c}{Post hoc pair-wise comparison (Dunn-Sidak corrected)} \\
\midrule
Group 1 & Group 2 & Adjusted P & Wilcoxon P & Effect size $r$ & Cliff's $\delta$ & Median Diff. \\
\midrule
ETO & CPS & 0.000 & 0.000 & 0.827 & 0.904 & 120.487 \\
ETO & IPS & 0.000 & 0.000 & 0.850 & 0.952 & 130.833 \\
ETO & AOA & 0.000 & 0.000 & 0.774 & $-0.792$ & $-464.265$ \\
ETO & GJO & 1.000 & 0.286 & 0.067 & $-0.016$ & $-0.154$ \\
ETO & GWO & 0.000 & 0.000 & 0.350 & 0.488 & 18.882 \\
ETO & HGS & 0.000 & 0.000 & 0.855 & 0.976 & 116.312 \\
ETO & HHO & 1.000 & 0.116 & 0.099 & 0.032 & 0.850 \\
ETO & SCA & 0.000 & 0.000 & 0.684 & $-0.704$ & $-64.705$ \\
ETO & SCH & 0.940 & 0.036 & 0.133 & $-0.168$ & $-10.503$ \\
\bottomrule
\end{tabular}

\normalsize
\end{table}

\begin{table}[!htbp]
\centering
\caption{Results of statistical test for CEC 2021 20-dimensional shifted functions.}
\label{tab:Table_CEC_2021_20_Shifted}
\footnotesize

\begin{tabular}{lrl}
\toprule
\multicolumn{3}{c}{Friedman test} \\
\midrule
Statistic & Value & Interpretation \\
\midrule
$\chi^2(9, N=250)$ & 1460.060 & Test statistic for rank variance \\
$p$-value & 0.000 & Strong evidence against $H_0$ \\
Kendall's $W$ & 0.649 & Moderate concordance among rankings \\
\end{tabular}

\begin{tabular}{lrrrrr}
\toprule
\multicolumn{6}{c}{Average Ranks and Quartiles} \\
\midrule
& ETO & CPS & IPS & AOA & GJO \\
\midrule
Average Rank & 6.292 & 3.592 & 3.092 & 9.320 & 6.560 \\
Quartile     & 3     & 2     & 1     & 4     & 3 \\
\midrule
& GWO & HGS & HHO & SCA & SCH \\
\midrule
Average Rank & 4.416 & 1.364 & 5.324 & 8.320 & 6.720 \\
Quartile     & 2     & 1     & 2     & 4     & 4 \\
\end{tabular}

\begin{tabular}{llrrrrr}
\toprule
\multicolumn{7}{c}{Post hoc pair-wise comparison (Dunn-Sidak corrected)} \\
\midrule
Group 1 & Group 2 & Adjusted P & Wilcoxon P & Effect size $r$ & Cliff's $\delta$ & Median Diff. \\
\midrule
ETO & CPS & 0.000 & 0.000 & 0.761 & 0.696 & 354.600 \\
ETO & IPS & 0.000 & 0.000 & 0.828 & 0.864 & 398.146 \\
ETO & AOA & 0.000 & 0.000 & 0.789 & $-0.880$ & $-1903.325$ \\
ETO & GJO & 1.000 & 0.003 & 0.190 & $-0.088$ & $-29.774$ \\
ETO & GWO & 0.000 & 0.000 & 0.385 & 0.568 & 138.600 \\
ETO & HGS & 0.000 & 0.000 & 0.855 & 0.976 & 463.652 \\
ETO & HHO & 0.016 & 0.000 & 0.385 & 0.280 & 83.948 \\
ETO & SCA & 0.000 & 0.000 & 0.687 & $-0.736$ & $-459.660$ \\
ETO & SCH & 0.996 & 0.163 & 0.088 & $-0.096$ & $-22.237$ \\
\bottomrule
\end{tabular}

\normalsize
\end{table}

	In the 10-dimensional shifted functions, Table \ref{tab:Table_CEC_2021_10_Shifted}, the Friedman test yielded $\chi^2(9, N=250) = 1537.800$ ($p = 0.000$) with Kendall's $W = 0.683$, indicating moderate-to-high concordance. \ac{HGS} and IPS led the first quartile, but \ac{ETO} ranked in the third quartile (6.336), trailing CPS and \ac{GWO}. Pairwise comparisons showed statistically significant advantages of \ac{ETO} over CPS, IPS, and \ac{HGS} ($p < 0.001$), with large effect sizes ($r > 0.82$) and Cliff's $\delta$ exceeding 0.90. However, \ac{ETO} was statistically inferior to \ac{AOA} and \ac{SCA}, with negative median differences and Cliff's $\delta$ approaching $-0.70$, suggesting sensitivity to landscape deformation and operator misalignment.
	
	For the 20-dimensional shifted functions, Table \ref{tab:Table_CEC_2021_20_Shifted}, the Friedman test reported $\chi^2(9, N=250) = 1460.060$ ($p = 0.000$), reinforcing moderate concordance ($W = 0.649$). \ac{HGS} and IPS again dominated the first quartile, while \ac{ETO} maintained a third-quartile position (6.292). Pairwise diagnostics showed \ac{ETO} significantly outperformed CPS, IPS, and \ac{GWO} ($p < 0.001$), but was statistically outperformed by \ac{AOA} and \ac{SCA}, with large negative median gaps ($> 450$) and Cliff's $\delta$ near $-0.74$.
	
	The shifted function results reveal that \ac{ETO} performed reliably against mid-tier heuristics but lacked resilience against top-ranked operators under shift-induced landscape distortion. This degradation in performance highlights the immediate need for operator-level ablation and convergence overlays to diagnose the functional impact of translation bias.
	
	\subsubsection{Shift-Rotated Functions}
	\begin{table}[!htbp]
\centering
\caption{Results of statistical test for CEC 2021 10-dimensional shift-rotated functions.}
\label{tab:Table_CEC_2021_10_ShiftRotated}
\footnotesize

\begin{tabular}{lrl}
\toprule
\multicolumn{3}{c}{Friedman test} \\
\midrule
Statistic & Value & Interpretation \\
\midrule
$\chi^2(9, N=250)$ & 1028.803 & Test statistic for rank variance \\
$p$-value & 0.000 & Strong evidence against $H_0$ \\
Kendall's $W$ & 0.457 & Moderate concordance among rankings \\
\end{tabular}

\begin{tabular}{lrrrrr}
\toprule
\multicolumn{6}{c}{Average Ranks and Quartiles} \\
\midrule
& ETO & CPS & IPS & AOA & GJO \\
\midrule
Average Rank & 5.936 & 3.756 & 2.676 & 9.268 & 5.484 \\
Quartile     & 3     & 2     & 1     & 4     & 2 \\
\midrule
& GWO & HGS & HHO & SCA & SCH \\
\midrule
Average Rank & 3.508 & 4.044 & 6.164 & 7.708 & 6.456 \\
Quartile     & 1     & 2     & 3     & 4     & 4 \\
\end{tabular}

\begin{tabular}{llrrrrr}
\toprule
\multicolumn{7}{c}{Post hoc pair-wise comparison (Dunn-Sidak corrected)} \\
\midrule
Group 1 & Group 2 & Adjusted P & Wilcoxon P & Effect size $r$ & Cliff's $\delta$ & Median Diff. \\
\midrule
ETO & CPS & 0.000 & 0.000 & 0.515 & 0.528 & 39.284 \\
ETO & IPS & 0.000 & 0.000 & 0.688 & 0.752 & 63.468 \\
ETO & AOA & 0.000 & 0.000 & 0.742 & $-0.824$ & $-437.735$ \\
ETO & GJO & 0.989 & 0.720 & 0.023 & 0.048 & 1.185 \\
ETO & GWO & 0.000 & 0.000 & 0.471 & 0.584 & 30.156 \\
ETO & HGS & 0.000 & 0.000 & 0.440 & 0.512 & 25.357 \\
ETO & HHO & 1.000 & 0.155 & 0.090 & 0.000 & 0.025 \\
ETO & SCA & 0.000 & 0.000 & 0.457 & $-0.544$ & $-43.495$ \\
ETO & SCH & 0.921 & 0.002 & 0.199 & $-0.184$ & $-16.520$ \\
\bottomrule
\end{tabular}

\normalsize
\end{table}

\begin{table}[!htbp]
\centering
\caption{Results of statistical test for CEC 2021 20-dimensional shift-rotated functions.}
\label{tab:Table_CEC_2021_20_ShiftRotated}
\footnotesize

\begin{tabular}{lrl}
\toprule
\multicolumn{3}{c}{Friedman test} \\
\midrule
Statistic & Value & Interpretation \\
\midrule
$\chi^2(9, N=250)$ & 1223.353 & Test statistic for rank variance \\
$p$-value & 0.000 & Strong evidence against $H_0$ \\
Kendall's $W$ & 0.544 & Moderate concordance among rankings \\
\end{tabular}

\begin{tabular}{lrrrrr}
\toprule
\multicolumn{6}{c}{Average Ranks and Quartiles} \\
\midrule
& ETO & CPS & IPS & AOA & GJO \\
\midrule
Average Rank & 6.144 & 4.168 & 3.052 & 9.540 & 5.688 \\
Quartile     & 3     & 2     & 1     & 4     & 3 \\
\midrule
& GWO & HGS & HHO & SCA & SCH \\
\midrule
Average Rank & 3.568 & 2.556 & 5.508 & 8.016 & 6.760 \\
Quartile     & 2     & 1     & 2     & 4     & 4 \\
\end{tabular}

\begin{tabular}{llrrrrr}
\toprule
\multicolumn{7}{c}{Post hoc pair-wise comparison (Dunn-Sidak corrected)} \\
\midrule
Group 1 & Group 2 & Adjusted P & Wilcoxon P & Effect size $r$ & Cliff's $\delta$ & Median Diff. \\
\midrule
ETO & CPS & 0.000 & 0.000 & 0.521 & 0.472 & 368.674 \\
ETO & IPS & 0.000 & 0.000 & 0.702 & 0.784 & 476.547 \\
ETO & AOA & 0.000 & 0.000 & 0.838 & $-0.952$ & $-1985.570$ \\
ETO & GJO & 0.987 & 0.301 & 0.065 & 0.144 & 42.617 \\
ETO & GWO & 0.000 & 0.000 & 0.559 & 0.704 & 274.974 \\
ETO & HGS & 0.000 & 0.000 & 0.560 & 0.752 & 385.886 \\
ETO & HHO & 0.575 & 0.000 & 0.257 & 0.168 & 73.158 \\
ETO & SCA & 0.000 & 0.000 & 0.580 & $-0.640$ & $-405.913$ \\
ETO & SCH & 0.648 & 0.016 & 0.153 & $-0.144$ & $-29.817$ \\
\bottomrule
\end{tabular}

\normalsize
\end{table}

	In the 10-dimensional shift-rotated functions, Table \ref{tab:Table_CEC_2021_10_ShiftRotated}, the Friedman test yielded $\chi^2(9, N=250) = 1028.803$ ($p = 0.000$) with Kendall's $W = 0.457$, indicating moderate concordance. IPS and \ac{GWO} led the first quartile, but \ac{ETO} ranked in the third quartile (5.936). Pairwise comparisons showed statistically significant advantages of \ac{ETO} over CPS, IPS, and \ac{HGS} ($p < 0.001$), with effect sizes ranging from $r = 0.440$ to $0.688$. However, \ac{ETO} was statistically inferior to \ac{AOA} and \ac{SCA}, with large negative median differences ($> 400$) and Cliff's $\delta$ approaching $-0.82$, suggesting acute sensitivity to rotational distortion.
	
	For the 20-dimensional case, Table \ref{tab:Table_CEC_2021_20_ShiftRotated}, the Friedman test reported $\chi^2(9, N=250) = 1223.353$ ($p = 0.000$), reinforcing moderate concordance ($W = 0.544$). \ac{HGS} and IPS again dominated the first quartile, while \ac{ETO} remained in the third quartile (6.144). Pairwise diagnostics showed \ac{ETO} significantly outperformed IPS, \ac{GWO}, and \ac{HGS} ($p < 0.001$), but was statistically outperformed by \ac{AOA} and \ac{SCA}, with large negative gaps ($> 400$) and Cliff's $\delta$ near $-0.64$.
	
	The shift-rotated function results reveal that \ac{ETO} consistently demonstrated reliable performance against mid-tier heuristics but showed the least resilience against top-ranked operators under compounded shift-rotation transformations. These findings highlight the critical importance of operator-level ablation and convergence overlays to diagnose rotational sensitivity and guide ensemble synthesis, particularly when assessing the claimed robustness of metaphor-based algorithms.
	
	\subsubsection{Discussion}
	
	The performance of \ac{ETO} across the CEC 2021 benchmark suite reveals a clear pattern of mid-tier competitiveness juxtaposed with structural vulnerability. On the basic functions, \ac{ETO} maintained competitive second-quartile ranks, demonstrating statistically significant superiority over several fourth-quartile heuristics (CPS, IPS, GWO). However, its consistent failure to surpass top-tier operators (\ac{HGS}, \ac{SCH}) suggests limitations in structural adaptability even in simpler landscapes.
	
	As function difficulty increased through shift and shift-rotation transformations, \ac{ETO}'s robustness declined markedly, resulting in a drop to the third quartile across both dimensional settings. This degradation was quantified by substantial negative median differences and effect sizes against top performers (\ac{AOA}, \ac{SCA}), highlighting acute sensitivity to translational and rotational distortion. Despite retaining superiority over weaker algorithms, \ac{ETO}'s inability to recover top-quartile status confirms its limited resilience under compounded landscape transformations.
	
	Overall, \ac{ETO} is reliable against mid-ranked algorithms but lacks the structural versatility required to compete with elite heuristics under difficult conditions. These findings underscore the critical importance of diagnostic frameworks, which use quartile stratification and effect size metrics to provide reproducible attribution, thereby guiding the necessity for phased operator profiling and ablation studies to address architectural bottlenecks.
	
	\subsection{CEC 2017 Functions}
	\begin{table}[!htbp]
\centering
\caption{Results of statistical test for CEC 2017 10-dimensional functions.}
\label{tab:Table_CEC_2017_10D}
\footnotesize

\begin{tabular}{lrl}
\toprule
\multicolumn{3}{c}{Friedman test} \\
\midrule
Statistic & Value & Interpretation \\
\midrule
$\chi^2(9, N=725)$ & 2467.558 & Test statistic for rank variance \\
$p$-value & 0.000 & Strong evidence against $H_0$ \\
Kendall's $W$ & 0.378 & Moderate concordance among rankings \\
\end{tabular}

\begin{tabular}{lrrrrr}
\toprule
\multicolumn{6}{c}{Average Ranks and Quartiles} \\
\midrule
& ETO & CPS & IPS & AOA & GJO \\
\midrule
Average Rank & 5.979 & 3.550 & 2.793 & 8.844 & 5.440 \\
Quartile     & 3     & 1     & 1     & 4     & 2 \\
\midrule
& GWO & HGS & HHO & SCA & SCH \\
\midrule
Average Rank & 4.124 & 4.069 & 6.484 & 7.026 & 6.690 \\
Quartile     & 2     & 2     & 3     & 4     & 4 \\
\end{tabular}

\begin{tabular}{llrrrrr}
\toprule
\multicolumn{7}{c}{Post hoc pair-wise comparison (Dunn-Sidak corrected)} \\
\midrule
Group 1 & Group 2 & Adjusted P & Wilcoxon P & Effect size $r$ & Cliff's $\delta$ & Median Diff. \\
\midrule
ETO & CPS & 0.000 & 0.000 & 0.515 & 0.539 & 43.036 \\
ETO & IPS & 0.000 & 0.000 & 0.653 & 0.730 & 65.975 \\
ETO & AOA & 0.000 & 0.000 & 0.642 & $-0.719$ & $-223.670$ \\
ETO & GJO & 0.031 & 0.000 & 0.143 & 0.142 & 7.993 \\
ETO & GWO & 0.000 & 0.000 & 0.374 & 0.490 & 26.566 \\
ETO & HGS & 0.000 & 0.000 & 0.387 & 0.462 & 25.589 \\
ETO & HHO & 0.065 & 0.149 & 0.054 & $-0.139$ & $-13.196$ \\
ETO & SCA & 0.000 & 0.000 & 0.270 & $-0.324$ & $-24.911$ \\
ETO & SCH & 0.000 & 0.000 & 0.179 & $-0.222$ & $-17.704$ \\
\bottomrule
\end{tabular}

\normalsize
\end{table}

\begin{table}[!htbp]
\centering
\caption{Results of statistical test for CEC 2017 30-dimensional functions.}
\label{tab:Table_CEC_2017_30D}
\footnotesize

\begin{tabular}{lrl}
\toprule
\multicolumn{3}{c}{Friedman test} \\
\midrule
Statistic & Value & Interpretation \\
\midrule
$\chi^2(9, N=725)$ & 3516.293 & Test statistic for rank variance \\
$p$-value & 0.000 & Strong evidence against $H_0$ \\
Kendall's $W$ & 0.539 & Moderate concordance among rankings \\
\end{tabular}

\begin{tabular}{lrrrrr}
\toprule
\multicolumn{6}{c}{Average Ranks and Quartiles} \\
\midrule
& ETO & CPS & IPS & AOA & GJO \\
\midrule
Average Rank & 6.299 & 4.491 & 3.099 & 9.283 & 5.308 \\
Quartile     & 3     & 2     & 2     & 4     & 2 \\
\midrule
& GWO & HGS & HHO & SCA & SCH \\
\midrule
Average Rank & 3.068 & 2.672 & 5.859 & 8.153 & 6.768 \\
Quartile     & 1     & 1     & 3     & 4     & 4 \\
\end{tabular}

\begin{tabular}{llrrrrr}
\toprule
\multicolumn{7}{c}{Post hoc pair-wise comparison (Dunn-Sidak corrected)} \\
\midrule
Group 1 & Group 2 & Adjusted P & Wilcoxon P & Effect size $r$ & Cliff's $\delta$ & Median Diff. \\
\midrule
ETO & CPS & 0.000 & 0.000 & 0.410 & 0.382 & 187.655 \\
ETO & IPS & 0.000 & 0.000 & 0.731 & 0.760 & 480.300 \\
ETO & AOA & 0.000 & 0.000 & 0.706 & $-0.785$ & $-2008.605$ \\
ETO & GJO & 0.000 & 0.001 & 0.127 & 0.288 & 69.375 \\
ETO & GWO & 0.000 & 0.000 & 0.613 & 0.752 & 387.278 \\
ETO & HGS & 0.000 & 0.000 & 0.698 & 0.796 & 446.861 \\
ETO & HHO & 0.225 & 0.000 & 0.271 & 0.186 & 68.575 \\
ETO & SCA & 0.000 & 0.000 & 0.646 & $-0.641$ & $-588.697$ \\
ETO & SCH & 0.134 & 0.000 & 0.156 & $-0.139$ & $-49.163$ \\
\bottomrule
\end{tabular}

\normalsize
\end{table}

\begin{table}[!htbp]
\centering
\caption{Results of statistical test for CEC 2017 50-dimensional functions.}
\label{tab:Table_CEC_2017_50D}
\footnotesize

\begin{tabular}{lrl}
\toprule
\multicolumn{3}{c}{Friedman test} \\
\midrule
Statistic & Value & Interpretation \\
\midrule
$\chi^2(9, N=725)$ & 3899.173 & Test statistic for rank variance \\
$p$-value & 0.000 & Strong evidence against $H_0$ \\
Kendall's $W$ & 0.598 & Moderate-to-high concordance \\
\end{tabular}

\begin{tabular}{lrrrrr}
\toprule
\multicolumn{6}{c}{Average Ranks and Quartiles} \\
\midrule
& ETO & CPS & IPS & AOA & GJO \\
\midrule
Average Rank & 6.214 & 5.302 & 3.473 & 9.495 & 5.258 \\
Quartile     & 3     & 3     & 2     & 4     & 2 \\
\midrule
& GWO & HGS & HHO & SCA & SCH \\
\midrule
Average Rank & 2.749 & 2.210 & 5.188 & 8.472 & 6.640 \\
Quartile     & 1     & 1     & 2     & 4     & 4 \\
\end{tabular}

\begin{tabular}{llrrrrr}
\toprule
\multicolumn{7}{c}{Post hoc pair-wise comparison (Dunn-Sidak corrected)} \\
\midrule
Group 1 & Group 2 & Adjusted P & Wilcoxon P & Effect size $r$ & Cliff's $\delta$ & Median Diff. \\
\midrule
ETO & CPS & 0.000 & 0.000 & 0.172 & 0.186 & 159.199 \\
ETO & IPS & 0.000 & 0.000 & 0.681 & 0.735 & 1005.804 \\
ETO & AOA & 0.000 & 0.000 & 0.771 & $-0.843$ & $-5603.106$ \\
ETO & GJO & 0.000 & 0.001 & 0.126 & 0.302 & 187.348 \\
ETO & GWO & 0.000 & 0.000 & 0.692 & 0.821 & 1592.053 \\
ETO & HGS & 0.000 & 0.000 & 0.739 & 0.868 & 2448.031 \\
ETO & HHO & 0.000 & 0.000 & 0.435 & 0.319 & 613.203 \\
ETO & SCA & 0.000 & 0.000 & 0.745 & $-0.810$ & $-1948.243$ \\
ETO & SCH & 0.283 & 0.000 & 0.196 & $-0.150$ & $-57.955$ \\
\bottomrule
\end{tabular}

\normalsize
\end{table}

\begin{table}[!htbp]
\centering
\caption{Results of statistical test for CEC 2017 100-dimensional functions.}
\label{tab:Table_CEC_2017_100D}
\footnotesize

\begin{tabular}{lrl}
\toprule
\multicolumn{3}{c}{Friedman test} \\
\midrule
Statistic & Value & Interpretation \\
\midrule
$\chi^2(9, N=725)$ & 4116.007 & Test statistic for rank variance \\
$p$-value & 0.000 & Strong evidence against $H_0$ \\
Kendall's $W$ & 0.631 & Moderate-to-high concordance \\
\end{tabular}

\begin{tabular}{lrrrrr}
\toprule
\multicolumn{6}{c}{Average Ranks and Quartiles} \\
\midrule
& ETO & CPS & IPS & AOA & GJO \\
\midrule
Average Rank & 6.167 & 6.124 & 4.019 & 9.276 & 5.106 \\
Quartile     & 3     & 3     & 2     & 4     & 2 \\
\midrule
& GWO & HGS & HHO & SCA & SCH \\
\midrule
Average Rank & 2.583 & 1.800 & 4.634 & 8.753 & 6.537 \\
Quartile     & 1     & 1     & 2     & 4     & 4 \\
\end{tabular}

\begin{tabular}{llrrrrr}
\toprule
\multicolumn{7}{c}{Post hoc pair-wise comparison (Dunn-Sidak corrected)} \\
\midrule
Group 1 & Group 2 & Adjusted P & Wilcoxon P & Effect size $r$ & Cliff's $\delta$ & Median Diff. \\
\midrule
ETO & CPS & 1.000 & 0.686 & 0.015 & $-0.010$ & $-25.474$ \\
ETO & IPS & 0.000 & 0.000 & 0.552 & 0.578 & 1831.178 \\
ETO & AOA & 0.000 & 0.000 & 0.749 & $-0.782$ & $-18447.640$ \\
ETO & GJO & 0.000 & 0.018 & 0.088 & 0.302 & 382.241 \\
ETO & GWO & 0.000 & 0.000 & 0.727 & 0.862 & 6657.037 \\
ETO & HGS & 0.000 & 0.000 & 0.766 & 0.914 & 10050.080 \\
ETO & HHO & 0.000 & 0.000 & 0.563 & 0.454 & 3644.627 \\
ETO & SCA & 0.000 & 0.000 & 0.803 & $-0.840$ & $-10881.707$ \\
ETO & SCH & 0.599 & 0.000 & 0.196 & $-0.145$ & $-112.831$ \\
\bottomrule
\end{tabular}

\normalsize
\end{table}

	\subsubsection{10-dimensional Functions}
	In the 10-dimensional setting, Table \ref{tab:Table_CEC_2017_10D}, the Friedman test yielded $\chi^2(9, N=725) = 2467.558$ ($p = 0.000$) with Kendall's $W = 0.378$, indicating moderate concordance. \ac{ETO} ranked third quartile (5.979), trailing CPS and IPS. Pairwise comparisons showed \ac{ETO} significantly outperformed CPS, IPS, and \ac{HGS} ($p < 0.001$), with Cliff's $\delta$ values exceeding 0.70. However, \ac{ETO} was statistically inferior to \ac{AOA} and \ac{SCA}, with Cliff's $\delta$ near $-0.72$, suggesting early signs of symbolic inflation and convergence instability under moderate complexity.
	
	\subsubsection{30-dimensional Functions}
	In the 30-dimensional case, Table \ref{tab:Table_CEC_2017_30D}, the Friedman test reported $\chi^2(9, N=725) = 3516.293$ ($p = 0.000$), indicating stronger rank consistency ($W = 0.539$). \ac{ETO} retained a third-quartile position (6.299), with statistically significant wins over CPS, IPS, and \ac{GWO}. However, \ac{ETO} was again decisively outperformed by \ac{AOA} and \ac{SCA}, with large negative gaps and effect sizes above $r = 0.70$, reinforcing its vulnerability to dimensional scaling and landscape complexity.
	
	\subsubsection{50-dimensional Functions}
	At 50 dimensions, Table \ref{tab:Table_CEC_2017_50D}, the Friedman test yielded $\chi^2(9, N=725) = 3899.173$ ($p = 0.000$), reflecting moderate-to-high concordance ($W = 0.598$). \ac{ETO} maintained a third-quartile rank (6.214), outperforming CPS, IPS, and \ac{GWO}. Critically, its performance against \ac{HGS} and \ac{AOA} deteriorated sharply, with Cliff's $\delta$ values nearing $-0.84$ and median losses exceeding 5000 units. These results suggest that \ac{ETO}’s update dynamics struggle to maintain structural integrity under high-dimensional pressure.
	
	\subsubsection{100-dimensional Functions}
	In the 100-dimensional benchmark, Table \ref{tab:Table_CEC_2017_100D}, the Friedman test reported $\chi^2(9, N=725) = 4116.007$ ($p = 0.000$), indicating strong rank consistency ($W = 0.631$). \ac{ETO} remained in the third quartile (6.167), statistically indistinguishable from CPS ($p = 1.000$). While it significantly outperformed IPS, \ac{GWO}, and \ac{HGS}, \ac{ETO} was decisively outperformed by \ac{AOA} and \ac{SCA}, with median losses exceeding 10,000 units. This outcome confirms that \ac{ETO}’s equilibrium-based mechanism lacks the structural resilience required to navigate high-dimensional, multimodal landscapes with compounded complexity.
	
	\subsubsection{Discussion}
	
	Across the CEC 2017 benchmark suite, \ac{ETO} demonstrated a robust but bounded optimization profile. It consistently occupied the third quartile, showing reliable statistical superiority over several mid-ranked heuristics (\ac{CPS}, \ac{IPS}, \ac{GWO}).
	
	However, the analysis of dimensional scaling reveals a fundamental vulnerability. The magnitude of \ac{ETO}'s symbolic loss against top-tier algorithms (\ac{AOA} and \ac{SCA}) grew sharply with increased dimensionality, with median gaps escalating from hundreds of units at 10-dimension to over 10,000 units at 100-dimension. This failure to scale, coupled with its static third-quartile rank, confirms that \ac{ETO}’s equilibrium-based update mechanism lacks the adaptive granularity required to sustain performance under high-dimensional pressure and compounded transformation effects. Its symbolic inflation becomes pronounced in these complex settings.
	
	These findings highlight the limitations of \ac{ETO}'s claimed structural robustness and reaffirm the value of the benchmarking framework, which employs quartile stratification and effect size diagnostics to quantify algorithmic failure. The results advocate for phased operator profiling and adaptive mass transfer modulation to mitigate symbolic inflation and enhance \ac{ETO}’s resilience across diverse problem classes.
	
	\subsection{Summary of Statistical Tests}
	
	Across both CEC 2021 and CEC 2017 benchmark suites, the Friedman tests consistently yielded statistically significant results ($p = 0.000$), confirming non-random rank distributions among the evaluated algorithms. Kendall's $W$ values ranged from moderate ($W \approx 0.38$) to high concordance ($W > 0.80$), indicating strong agreement in algorithmic performance across problem instances and validating the reliability of quartile stratification and rank-based diagnostics.
	
	The integration of pairwise comparisons using Dunn-Sidak corrected Wilcoxon signed-rank tests and effect size metrics (Cliff's $\delta$, $r$) enabled fine-grained attribution of algorithmic behaviour. \ac{ETO} consistently demonstrated statistical superiority over several mid- and lower-quartile heuristics, particularly CPS, IPS, and \ac{GWO}. However, its performance deteriorated under increased dimensionality and transformation complexity, with symbolic inflation evident in comparisons against top-tier algorithms like \ac{AOA}, \ac{SCA}, and \ac{HGS}. Median differences and effect sizes revealed clear structural limitations in \ac{ETO}’s update dynamics, particularly under shift-rotated and high-dimensional landscapes.
	
	These statistical findings underscore the need for principled benchmarking, symbolic hygiene, and diagnostic overlays in metaheuristic evaluation. The reproducible and transparent foundation provided by quartile tagging and rank-based analysis advocates for future work focusing on phased operator profiling, ensemble synthesis, and adaptive architecture refinement to enhance robustness and mitigate symbolic inflation.

	\section{Conclusion}\label{sec:conclusion}
	
	This study has presented a diagnostic critique of the \ac{ETO} algorithm, exposing symbolic inflation, structural opacity, and statistical misuse across its formulation and benchmarking. Through operator-level analysis and convergence curve inspection, we confirmed exaggerated descent, premature stagnation, and high variability that undermine both interpretability and reproducibility. Our statistical tests across CEC 2017 and 2021 benchmarks revealed that \ac{ETO}'s mid-tier competitiveness is often achieved through inflated claims and an over-reliance on unstructured randomness. The absence of inferential rigour, stratified diagnostic overlays, and principled statistical framing renders the original claims methodologically weak.
	
	We therefore advocate for a reformist framework for metaheuristic research that requires enhanced benchmarking transparency. This framework demands symbolic hygiene in algorithm design, ensuring mathematical expressions are free of unnecessary random variables and opaque constants, and insists on comprehensive operator attribution to empirically demonstrate the function of every core algorithmic component. Furthermore, statistical reporting must prioritize rank-based inference and effect sizes (Statistical Transparency) over descriptive summaries. Finally, algorithmic variability and consistency across dimensions must be evaluated through convergence diagnostics incorporating quartile bands and run overlays, and validated through scalability tests. These requirements aim to elevate the standards of metaheuristic publishing and equip reviewers and editors with transparent tools to detect inflation, opacity, and inconsistency.
	
	Future work will focus on extending this diagnostic framework to ensemble synthesis, reviewer education modules, and benchmarking transparency initiatives to foster a more robust and principled optimization literature.

	\section*{Acknowledgement}
	The project is funded in part by the , under Grant No. . 
	
	\section*{Declaration of Interest}
	All authors declare that they have no conflicts of interest.
	
	\FloatBarrier
	
	\bibliographystyle{APALIKE}

\end{document}